\documentclass[10pt,twocolumn,letterpaper]{article}

\usepackage[pagenumbers]{cvpr} %

\usepackage{graphicx}
\usepackage{amsmath}
\usepackage{amssymb}
\usepackage{booktabs}
\usepackage{soul}
\usepackage{nicefrac}
\usepackage{pifont}
\usepackage{footmisc}
\usepackage{multirow}
\usepackage{array}

\usepackage[dvipsnames,table,xcdraw]{xcolor}
\usepackage{adjustbox}

\usepackage{caption}
\usepackage{subcaption}
\usepackage{diagbox}

\usepackage{lipsum}   %
\usepackage[accsupp]{axessibility}

\usepackage{tikz}
\usetikzlibrary{arrows,shapes,calc,matrix,fit,backgrounds}
\usepackage{pgfplots}
\usepackage{pgfplotstable}
\pgfplotsset{compat=1.9}
\usepgfplotslibrary{fillbetween}

\usepackage{xstring}

\usepgfplotslibrary{external}

\IfBeginWith*{\jobname}{fig/extern/}{\finalcopy}{}

\tikzset{every mark/.append style={solid}}
\pgfplotsset{%
	grid=both, width=\columnwidth, try min ticks=5,
	every axis plot/.append style={thick,mark=none,mark size=1.2,tension=0.18},
	legend cell align=left, legend style={fill opacity=0.8},
}

\pgfplotsset{
	dash/.style={mark=o,dashed,opacity=0.7},
	dott/.style={mark=o,dotted,opacity=0.7},
}

\definecolor{cvprblue}{rgb}{0.21,0.49,0.74}
\usepackage[pagebackref,breaklinks,colorlinks,citecolor=cvprblue]{hyperref}

\def\paperID{4977} %
\def\confName{CVPR}
\def\confYear{2024}

\begin{document}

\title{Label Propagation for Zero-shot Classification with Vision-Language Models}

\author{
    Vladan Stojnić$^1$
    \and
    Yannis Kalantidis$^2$
    \and
    Giorgos Tolias$^1$ \and \\
    $^1$~{VRG, FEE, Czech Technical University in Prague} \hspace{1cm} $^2$~{NAVER LABS Europe}
}

\maketitle

\newcommand{\mypartight}[1]{\noindent {\bf #1}}
\newcommand{\myparagraph}[1]{\vspace{3pt}\noindent\textbf{#1}\xspace}

\newcommand{\alert}[1]{{\color{red}{#1}}}
\newcommand{\gt}[1]{{\color{purple}{GT: #1}}}
\newcommand{\gtt}[1]{{\color{purple}{#1}}}
\newcommand{\gtr}[2]{{\color{purple}\st{#1} {#2}}}
\newcommand{\vsc}[1]{{\color{blue}{VS: #1}}}
\newcommand{\yk}[1]{{\color{OliveGreen}{Y: #1}}}
\newcommand{\ykt}[1]{{\color{OliveGreen}{#1}}}
\newcommand{\cready}[1]{{\color{RoyalPurple}{#1}}}

\newcommand{\gray}[1]{{\color{gray}{#1}}}

\newcommand{\gain}[1]{\textbf{\color{OliveGreen}{{\scriptsize$\uparrow$}{\scriptsize #1}}}}
\newcommand{\loss}[1]{\textbf{\color{red}{{\scriptsize$\downarrow$}{\scriptsize #1}}}}

\newcommand{\rvi}[1]{{\color{NavyBlue}{\textbf{#1}}}}
\newcommand{\rvii}[1]{{\color{OliveGreen}{\textbf{#1}}}}
\newcommand{\rviii}[1]{{\color{BrickRed}{\textbf{#1}}}}

\newcommand{\rone}[0]{{\rvi{R-wyHD}\xspace}}
\newcommand{\rtwo}[0]{{\rvii{R-Gt7B}\xspace}}
\newcommand{\rthree}[0]{{\rviii{R-YJqj}\xspace}}

\newcommand{\question}[1]{\noindent\emph{\textbf{#1}}}

\newcommand{\questionone}[1]{\noindent\emph{\textbf{\rvi{#1}}}}
\newcommand{\questiontwo}[1]{\noindent\emph{\textbf{\rvii{#1}}}}
\newcommand{\questionthree}[1]{\noindent\emph{\textbf{\rviii{#1}}}}

\def\roxf{$\mathcal{R}$Oxford\xspace}
\def\rox{$\mathcal{R}$Oxf\xspace}
\def\ro{$\mathcal{R}$O\xspace}
\def\rpar{$\mathcal{R}$Paris\xspace}
\def\rpa{$\mathcal{R}$Par\xspace}
\def\rp{$\mathcal{R}$P\xspace}
\def\rdis{$\mathcal{R}$1M\xspace}

\newcommand\resnet[3]{\ensuremath{\prescript{#2}{}{\mathtt{R}}{#1}_{\scriptscriptstyle #3}}\xspace}

\newcommand{\clipc}{CornflowerBlue}
\newcommand{\oursc}{MidnightBlue}
\newcommand{\ourssparsec}{MidnightBlue!50!Black}

\newcommand{\inmapc}{RedOrange}
\newcommand{\oursinmapc}{BrickRed}
\newcommand{\oursinmapsparsec}{BrickRed!50!Black}

\newcommand{\clipdnc}{OliveGreen}
\newcommand{\tptc}{DarkOrchid}
\newcommand{\ours}{ZLaP\xspace} %
\newcommand{\ourssparse}{\text{ZLaP}\ensuremath{^{\ast}}\xspace} %
\newcommand{\oursinmap}{iZLaP\xspace} %

\newcommand{\stddev}[1]{\scriptsize{$\pm#1$}}

\newcommand{\diffup}[1]{{\color{OliveGreen}{($\uparrow$ #1)}}}
\newcommand{\diffdown}[1]{{\color{BrickRed}{($\downarrow$ #1)}}}

\newcommand{\comment} [1]{{\color{orange} \Comment     #1}} %

\def\nmsp{\hspace{-6pt}}
\def\nssp{\hspace{-3pt}}
\def\nxssp{\hspace{-1pt}}
\def\zsp{\hspace{0pt}}
\def\xssp{\hspace{1pt}}
\def\ssp{\hspace{3pt}}
\def\msp{\hspace{6pt}}
\def\lsp{\hspace{12pt}}
\def\xlsp{\hspace{20pt}}

\newcommand{\head}[1]{{\smallskip\noindent\bf #1}}
\newcommand{\equ}[1]{(\ref{equ:#1})\xspace}

\def\Linv{\ensuremath{L_{\text{inv}}}\xspace}

\newcommand{\nn}[1]{\ensuremath{\text{NN}_{#1}}\xspace}
\newcommand{\knn}{\ensuremath{\text{kNN}}\xspace}
\def\l1{\ensuremath{\ell_1}\xspace}
\def\l2{\ensuremath{\ell_2}\xspace}

\newcommand{\tran}{^\top}
\newcommand{\mtran}{^{-\top}}
\newcommand{\zcol}{\mathbf{0}}
\newcommand{\zrow}{\zcol\tran}

\newcommand{\ind}{\mathds{1}}
\newcommand{\expect}{\mathbb{E}}
\newcommand{\nat}{\mathbb{N}}
\newcommand{\zahl}{\mathbb{Z}}
\newcommand{\real}{\mathbb{R}}
\newcommand{\proj}{\mathbb{P}}
\newcommand{\prob}{\mathbf{Pr}}

\newcommand{\mif}{\textrm{if }}
\newcommand{\other}{\textrm{otherwise}}
\newcommand{\minimize}{\textrm{minimize }}
\newcommand{\maximize}{\textrm{maximize }}

\newcommand{\id}{\operatorname{id}}
\newcommand{\const}{\operatorname{const}}
\newcommand{\sgn}{\operatorname{sgn}}
\newcommand{\var}{\operatorname{Var}}
\newcommand{\mean}{\operatorname{mean}}
\newcommand{\trace}{\operatorname{tr}}
\newcommand{\diag}{\operatorname{diag}}
\newcommand{\vect}{\operatorname{vec}}
\newcommand{\cov}{\operatorname{cov}}
\newcommand{\argmax}{\operatorname{argmax}}

\newcommand{\softmax}{\operatorname{softmax}}
\newcommand{\clip}{\operatorname{clip}}

\newcommand{\defn}{\mathrel{:=}}
\newcommand{\peq}{\mathrel{+\!=}}
\newcommand{\meq}{\mathrel{-\!=}}

\newcommand{\floor}[1]{\left\lfloor{#1}\right\rfloor}
\newcommand{\ceil}[1]{\left\lceil{#1}\right\rceil}
\newcommand{\inner}[1]{\left\langle{#1}\right\rangle}
\newcommand{\norm}[1]{\left\|{#1}\right\|}
\newcommand{\frob}[1]{\norm{#1}_F}
\newcommand{\card}[1]{\left|{#1}\right|\xspace}
\newcommand{\diff}{\mathrm{d}}
\newcommand{\der}[3][]{\frac{d^{#1}#2}{d#3^{#1}}}
\newcommand{\pder}[3][]{\frac{\partial^{#1}{#2}}{\partial{#3^{#1}}}}
\newcommand{\ipder}[3][]{\partial^{#1}{#2}/\partial{#3^{#1}}}
\newcommand{\dder}[3]{\frac{\partial^2{#1}}{\partial{#2}\partial{#3}}}

\newcommand{\wb}[1]{\overline{#1}}
\newcommand{\wt}[1]{\widetilde{#1}}

\newcommand{\cA}{\mathcal{A}}
\newcommand{\cB}{\mathcal{B}}
\newcommand{\cC}{\mathcal{C}}
\newcommand{\cD}{\mathcal{D}}
\newcommand{\cE}{\mathcal{E}}
\newcommand{\cF}{\mathcal{F}}
\newcommand{\cG}{\mathcal{G}}
\newcommand{\cH}{\mathcal{H}}
\newcommand{\cI}{\mathcal{I}}
\newcommand{\cJ}{\mathcal{J}}
\newcommand{\cK}{\mathcal{K}}
\newcommand{\cL}{\mathcal{L}}
\newcommand{\cM}{\mathcal{M}}
\newcommand{\cN}{\mathcal{N}}
\newcommand{\cO}{\mathcal{O}}
\newcommand{\cP}{\mathcal{P}}
\newcommand{\cQ}{\mathcal{Q}}
\newcommand{\cR}{\mathcal{R}}
\newcommand{\cS}{\mathcal{S}}
\newcommand{\cT}{\mathcal{T}}
\newcommand{\cU}{\mathcal{U}}
\newcommand{\cV}{\mathcal{V}}
\newcommand{\cW}{\mathcal{W}}
\newcommand{\cX}{\mathcal{X}}
\newcommand{\cY}{\mathcal{Y}}
\newcommand{\cZ}{\mathcal{Z}}

\newcommand{\vA}{\mathbf{A}}
\newcommand{\vB}{\mathbf{B}}
\newcommand{\vC}{\mathbf{C}}
\newcommand{\vD}{\mathbf{D}}
\newcommand{\vE}{\mathbf{E}}
\newcommand{\vF}{\mathbf{F}}
\newcommand{\vG}{\mathbf{G}}
\newcommand{\vH}{\mathbf{H}}
\newcommand{\vI}{\mathbf{I}}
\newcommand{\vJ}{\mathbf{J}}
\newcommand{\vK}{\mathbf{K}}
\newcommand{\vL}{\mathbf{L}}
\newcommand{\vM}{\mathbf{M}}
\newcommand{\vN}{\mathbf{N}}
\newcommand{\vO}{\mathbf{O}}
\newcommand{\vP}{\mathbf{P}}
\newcommand{\vQ}{\mathbf{Q}}
\newcommand{\vR}{\mathbf{R}}
\newcommand{\vS}{\mathbf{S}}
\newcommand{\vT}{\mathbf{T}}
\newcommand{\vU}{\mathbf{U}}
\newcommand{\vV}{\mathbf{V}}
\newcommand{\vW}{\mathbf{W}}
\newcommand{\vX}{\mathbf{X}}
\newcommand{\vY}{\mathbf{Y}}
\newcommand{\vZ}{\mathbf{Z}}

\newcommand{\va}{\mathbf{a}}
\newcommand{\vb}{\mathbf{b}}
\newcommand{\vc}{\mathbf{c}}
\newcommand{\vd}{\mathbf{d}}
\newcommand{\ve}{\mathbf{e}}
\newcommand{\vf}{\mathbf{f}}
\newcommand{\vg}{\mathbf{g}}
\newcommand{\vh}{\mathbf{h}}
\newcommand{\vi}{\mathbf{i}}
\newcommand{\vj}{\mathbf{j}}
\newcommand{\vk}{\mathbf{k}}
\newcommand{\vl}{\mathbf{l}}
\newcommand{\vm}{\mathbf{m}}
\newcommand{\vn}{\mathbf{n}}
\newcommand{\vo}{\mathbf{o}}
\newcommand{\vp}{\mathbf{p}}
\newcommand{\vq}{\mathbf{q}}
\newcommand{\vr}{\mathbf{r}}
\newcommand{\Vs}{\mathbf{s}}
\newcommand{\vt}{\mathbf{t}}
\newcommand{\vu}{\mathbf{u}}
\newcommand{\vv}{\mathbf{v}}
\newcommand{\vw}{\mathbf{w}}
\newcommand{\vx}{\mathbf{x}}
\newcommand{\vy}{\mathbf{y}}
\newcommand{\vz}{\mathbf{z}}

\newcommand{\vone}{\mathbf{1}}
\newcommand{\vzero}{\mathbf{0}}

\newcommand{\valpha}{{\boldsymbol{\alpha}}}
\newcommand{\vbeta}{{\boldsymbol{\beta}}}
\newcommand{\vgamma}{{\boldsymbol{\gamma}}}
\newcommand{\vdelta}{{\boldsymbol{\delta}}}
\newcommand{\vepsilon}{{\boldsymbol{\epsilon}}}
\newcommand{\vzeta}{{\boldsymbol{\zeta}}}
\newcommand{\veta}{{\boldsymbol{\eta}}}
\newcommand{\vtheta}{{\boldsymbol{\theta}}}
\newcommand{\viota}{{\boldsymbol{\iota}}}
\newcommand{\vkappa}{{\boldsymbol{\kappa}}}
\newcommand{\vlambda}{{\boldsymbol{\lambda}}}
\newcommand{\vmu}{{\boldsymbol{\mu}}}
\newcommand{\vnu}{{\boldsymbol{\nu}}}
\newcommand{\vxi}{{\boldsymbol{\xi}}}
\newcommand{\vomikron}{{\boldsymbol{\omikron}}}
\newcommand{\vpi}{{\boldsymbol{\pi}}}
\newcommand{\vrho}{{\boldsymbol{\rho}}}
\newcommand{\vsigma}{{\boldsymbol{\sigma}}}
\newcommand{\vtau}{{\boldsymbol{\tau}}}
\newcommand{\vupsilon}{{\boldsymbol{\upsilon}}}
\newcommand{\vphi}{{\boldsymbol{\phi}}}
\newcommand{\vchi}{{\boldsymbol{\chi}}}
\newcommand{\vpsi}{{\boldsymbol{\psi}}}
\newcommand{\vomega}{{\boldsymbol{\omega}}}

\newcommand{\rLambda}{\mathrm{\Lambda}}
\newcommand{\rSigma}{\mathrm{\Sigma}}

\makeatletter
\DeclareRobustCommand\onedot{\futurelet\@let@token\@onedot}
\def\@onedot{\ifx\@let@token.\else.\null\fi\xspace}
\def\eg{\emph{e.g}\onedot} \def\Eg{\emph{E.g}\onedot}
\def\ie{\emph{i.e}\onedot} \def\Ie{\emph{I.e}\onedot}
\def\vs{\emph{vs\onedot}}
\def\cf{\emph{cf}\onedot} \def\Cf{\emph{C.f}\onedot}
\def\etc{\emph{etc}\onedot} \def\vs{\emph{vs}\onedot}
\def\wrt{w.r.t\onedot} \def\dof{d.o.f\onedot}
\def\etal{\emph{et al}\onedot}
\makeatother

\begin{abstract}

Vision-Language Models (VLMs) have demonstrated impressive performance on zero-shot classification, \ie classification when provided merely with a list of class names. 
In this paper, we tackle the case of zero-shot classification in the presence of unlabeled data.
We leverage the graph structure of the unlabeled data and introduce \textbf{\ours}, a method based on label propagation (LP) that utilizes geodesic distances for classification. We tailor LP to graphs containing both text and image features and further propose an efficient method for performing inductive inference based on a dual solution and a sparsification step. 
We perform extensive experiments to evaluate the effectiveness of our method on 14 common datasets and show that \ours outperforms the latest related works. Code: \url{https://github.com/vladan-stojnic/ZLaP}

\end{abstract}

\vspace{5pt}

\section{Introduction}
\label{sec:intro}

Vision-Language Models (VLMs) have demonstrated impressive performance on a variety of computer vision tasks. They are usually trained on large datasets of image-text pairs and contain visual and textual encoders that map to a common feature space. Visual encoders from such models have been shown to produce strong visual representations for perception tasks~\cite{rkh+21}. Given labeled data from a downstream dataset, one can fine-tune the model or learn classifiers and achieve really high classification accuracy.

Besides using the visual encoder in isolation, the joint text and visual encoder feature space of VLMs enables us to define text-based ``classifiers'', \eg using the class names as textual prompts. This means that we only need a list of class names to perform \emph{zero-shot} classification for a target dataset, \ie without access to any labeled images.
Although utilizing priors or devising better textual prompts can improve zero-shot performance~\cite{mv23,plf22,gzq+23,zrl+23}, here, we are interested in the case where we further have access to \textit{unlabeled} data.
Our goal is to find the best way of utilizing such data for zero-shot classification.

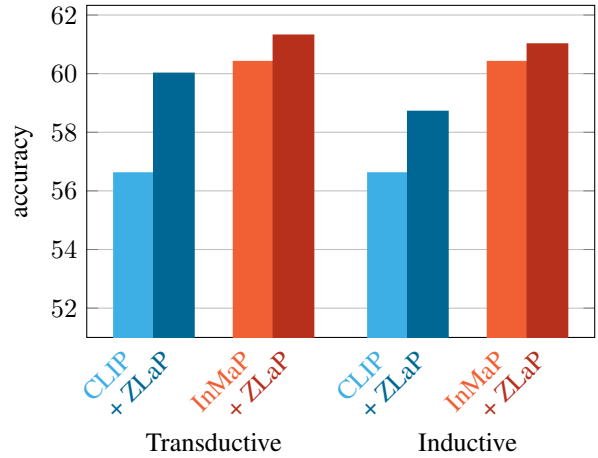
\begin{figure}[t]
  \centering
  \definecolor{rred}{HTML}{C0504D}
\definecolor{ggreen}{HTML}{9BBB59}
\definecolor{ppurple}{HTML}{9F4C7C}

\adjustbox{max width=\linewidth}
{
\begin{tikzpicture}
    \begin{axis}[
        width  = \linewidth,
        height = 6cm,
        major x tick style = transparent,
        ybar=0.1\pgflinewidth,
        bar width=15pt,
        ylabel = {accuracy},
        symbolic x coords={Transductive, Inductive},
        xtick = data,
        scaled y ticks = false,
        enlarge x limits=0.5,
        ymin=51,
        x tick label style={yshift=-10mm},
        scatter/position=absolute,
        grid=none,
        ymajorgrids=true,
        node near coords style={
            at={(axis cs:\pgfkeysvalueof{/data point/x},\pgfkeysvalueof{/pgfplots/ymin})},
            yshift={-\pgfkeysvalueof{/pgfplots/major tick length}},
           anchor=east,
           rotate=45,
        },
    ]
        \addplot[style={\clipc,fill=\clipc,mark=none}, 
        nodes near coords=CLIP] 
            coordinates {(Transductive, 56.6) (Inductive, 56.6)};

        \addplot[style={\oursc,fill=\oursc,mark=none},
        nodes near coords=  + \ours]
             coordinates {(Transductive,60.0) (Inductive,58.7)};

        \addplot[style={\oursc,fill=\oursc,mark=none}]
             coordinates {};

        \addplot[style={\inmapc,fill=\inmapc,mark=none},
        nodes near coords=InMaP] 
             coordinates {(Transductive,60.4) (Inductive,60.4)};

        \addplot[style={\oursinmapc,fill=\oursinmapc,mark=none},
        nodes near coords=  + \ours]
             coordinates {(Transductive,61.3) (Inductive,61.0)};
        
    \end{axis}
\end{tikzpicture}
}
  \caption{\textbf{Zero-shot classification performance over 14 datasets} using the proposed \ours classifier over CLIP~\cite{rkh+21}, as well as over the (concurrent) InMaP~\cite{qxh23} approach. 
  Our method offers performance gains for both transductive (left) and inductive (right) inference. Average accuracy over 14 common datasets is reported.}
  \label{fig:teaser}
\end{figure}

In this paper, we leverage the inherent structure of the unlabeled data represented by a proximity graph and apply \emph{label propagation} (LP) between the text-based classifiers and unlabeled images to derive geodesic distances we then use for classification.
We tailor LP to VLMs and graphs containing both text and image features, and show that without proper handling of the bimodality, vanilla application of LP fails dramatically.
We introduce \ours, a novel 
classification method based on label propagation
that can perform both transductive and inductive inference. 
We perform the former with the standard (primal) solution of LP for classification 
and devise a more efficient dual solution for the latter. 
Our method is not only highly effective but also efficient, making LP a more attractive inductive classifier in terms of complexity.

We implement our methods using 
publicly available VLMs
as feature encoders, 
primarily the ResNet and ViT CLIP~\cite{rkh+21} models,
and perform extensive experiments to evaluate the effectiveness of our method on 14 common datasets. We show that we are able to achieve top performance on two zero-shot inference setups, \ie inductive and transductive inference. 
\Cref{fig:teaser} summarizes our gains over 14 datasets on both setups when applying our LP-powered classifiers on top of CLIP, as well as after incorporating the class proxies from the recent InMaP~\cite{qxh23} zero-shot approach.

It is worth highlighting that \ours is a non-parametric method, \ie it does not involve a learning step. In fact, our approach does not even require access to the VLM model weights and can therefore be used to improve the zero-shot performance of a black-box model even, \eg provided only via an API.
In summary, our contributions are as follows.
\begin{itemize}
 	\setlength\itemsep{3pt} %
    \item We tailor label propagation to VLMs 
    and zero-shot classification over bi-modal graphs, proposing per modality neighbor search and balancing of contributions.
    \item We propose an efficient way for performing inductive inference with label propagation via a dual solution and through sparsification. This not only improves the test-time efficiency of our method but also performance.
    \item We complement our method with the class proxies presented in concurrent work~\cite{qxh23} and achieve state of the art results for zero-shot on 14 common datasets.
\end{itemize}
\section{Related work}
\label{sec:related}
In this section, we discuss related works that improve the already impressive zero-shot classification performance of vision-language models~\cite{rkh+21,jyx+21,cbw+23} even further. This is achieved by devising better distance metrics, utilizing external knowledge to learn more expressive textual prompts, or by leveraging synthetic and unlabeled data.

\paragraph{Improved distance metrics.} 
Zero-shot classification can be improved by devising a better distance metric between image and text representations~\cite{gzq+23,zrl+23}. 
CALIP~\cite{gzq+23} uses a parameter-free attention mechanism and a local patch representation, \ie instead of global representations, to improve the estimation of class-to-image similarity. 
CLIP-DN~\cite{zrl+23} improves the test-time similarity estimation by alignment with the similarity used during contrastive pre-training of VLMs. To achieve this, the method assumes access to unlabeled data from the target distribution. 
TPT~\cite{snh+22} optimizes textual prompts via a consistency objective across test image augmentations. 
Our method can be considered a part of this line of work, as label propagation is a similarity measure in a geodesic space instead of the Euclidean space.

\paragraph{Improved textual prompts using language models.}
CLIP~\cite{rkh+21} uses hand-crafted prompts that are specialized for each domain. 
Instead of using hand-crafted prompts, generating them with large language models (LLMs) is shown to be promising~\cite{mv23,plf22}. 
VisDesc~\cite{mv23} and CuPL~\cite{plf22} query LLMs to generate diverse descriptions of all classes, while WaffleCLIP~\cite{rkk+23} operates on top of VisDesc to systematically analyze which parts of the generated prompts are the most important. 
Instead of generating class descriptions, CHiLS~\cite{nml+23} targets diversifying the set of classes, by generating sub-classes per class, either through an existing class hierarchy or by querying an LLM. 
It then performs zero-shot classification using sub-classes and linking them to the parent class. 
We show that methods improving the textual prompts are complementary to our approach.

\paragraph{Synthetic data.}
Recent methods~\cite{hsy+23,uga23,zhl+23} demonstrate that the use of synthetic data is beneficial for zero-shot classification. 
CLIP+SYN~\cite{hsy+23} uses a stable-diffusion-based model to generate synthetic images using class names and uses them to train a linear classifier, initialized by the VLM class representations. 
SuS-X~\cite{uga23} considers a similar approach, but relies on a non-parametric classifier.
CaFO~\cite{zhl+23} follows the same path, but additionally includes text prompts generated by LLMs. %

\paragraph{External datasets.}
Besides the use of synthetic data, SuS-X proposes a variant that operates on an extensive unlabeled image dataset (LAION-5B~\cite{sbv+22}). This dataset encompasses  a distribution that is a super set of the target one.
The method generates pseudo-labels using the zero-shot approach which are then incorporated within the non-parametric classifier. NeuralPriming~\cite{wrf+23} additionally assumes that images have captions, which are used to improve the pseudo-labeling.

\paragraph{Unlabeled images from the target distribution.}
Another line of research~\cite{hzx+23,qxh23,hcw22,pbj+22} propose operating on unlabeled datasets from the target distribution. 
The main ingredient of all these methods is the prediction of pseudo-labels for unlabeled examples, that are later used for further processing.
UPL~\cite{hcw22} optimizes learnable text prompts based on the pseudo-labels.
SVL-Adapter~\cite{pbj+22} first trains a self-supervised model on unlabeled data, and then an adapter module to align its outputs to the pseudo-labels. 
ReCLIP~\cite{hzx+23} performs transductive label propagation to obtain the pseudo-labels and uses them to fine-tune the VLM visual and textual encoders. 
In contrast to that, we do not fine-tune the model, which we may not have access to, and efficiently use label propagation for inductive inference too.
InMaP~\cite{qxh23} is a concurrent work that uses pseudo-labels to update the class representations such that they are now closer to image representations. 
We show in the experiments that this approach is complementary to ours.
In contrast to all those methods, we do not explicitly require pseudo-label prediction, but rather capture interactions between all unlabeled examples through a proximity graph and label propagation.

\section{Method}
\label{sec:method}
We first define the task of zero-shot classification with access to unlabeled examples, present label propagation with our contributions
and then present the proposed approach for zero-shot classification using unlabeled examples.

\subsection{Problem formulation}
Vision-language models consist of an image encoder $f: \cI \rightarrow \real^{d}$ and a text encoder $g: \cT \rightarrow \real^{d}$, where $\cI$ and $\cT$ represent the space of images and text, respectively. 
We consider the outputs of these encoders to be \l2-normalized. 

Let $\cC$ denote a set of known classes with associated class names $\left\{l_1,...,l_C\right\}$ and $\cP = \left\{p_1,...,p_P\right\}$ a set of prompt templates. Each prompt is combined with a class name to produce a textual description of the class, \ie $p_i(l_c)$ for the $i$-th template used for class $c$. Class representations $\vw_c = \nicefrac{1}{P}\sum_{i=1}^{P}g(p_i(l_c))$ are obtained using the VLM. Then, for a test image $u$, we extract 
representation $\vu = f(u)$, and perform zero-shot classification by $\argmax_c \vu^{T}\vw_c$.

We further assume access to a set $\cU$ of $M$ unlabeled images.
Let $\{\vu_1, \ldots, \vu_M\}$ denote the representations of the unlabeled images.

In this work, we assume no direct access to the VML model weights, \ie VLM training is neither possible nor desired.
This allows to consider the underlying VLM as a black box, possibly only available through an API that generates features.
We consider two inference setups:

\noindent\textbf{Inductive inference:}
We 
consider an \emph{inductive} setup, where we need to construct a classifier that can operate on new examples. This classifier should take advantage of the unlabeled examples in $\cU$.

\noindent\textbf{Transductive inference:}    
We consider $\cU$ to be the test set, \ie all test examples are jointly provided in advance.
Prediction for test example $\vu_i$ may therefore depend on the representations and predictions of all other test examples. 
In this \emph{transductive} setup the models are not required to provide predictions for any example that is not in $\cU$.

\subsection{Label propagation (LP)}
\label{sec:lp}
Let $\left\{\vx_1,...,\vx_N\right\}$, with $\vx_i \in \real^{d}$, be a set of features for $N$ examples.
Each feature represents a graph node.
We construct an adjacency matrix $S \in \real^{N \times N}$ with zero diagonal,
and $s_{ij}$ equal to $\vx_i^\top \vx_j$ if  $\vx_j$ is in the $k$-nearest neighbors of $\vx_i$ (denoted by $\vx_j \in \knn(\vx_i))$, and 0 otherwise.
We obtain a symmetric adjacency matrix 
by $\bar{S}=S+S^\top$, and its symmetrically normalized version by $\hat{S} = D^{-\frac{1}{2}}\bar{S}D^{-\frac{1}{2}}$, where $D = \diag(\bar{S}\vone_N)$ is the degree matrix, and $\vone_N$ is the all-ones $N$-dimensional vector. 
We assume the first $C$ examples, and the corresponding nodes, to be labeled among $C$ classes; each class is assigned to a single node\footnote{The theoretical part described in this section holds for the case of more labeled examples per class too. We consider this specific  case for simplicity of the presentation and because it corresponds to the task of zero-shot classification with unlabeled examples.}.

\paragraph{Transductive inference.} Label propagation~\cite{zbl+03} is originally proposed for the transductive inference setup; we need to predict labels for the unlabeled nodes of the graph.
Given the normalized adjacency matrix $\hat{S}$, label propagation is an iterative process given by
\begin{equation}
    \hat{\vy}^{(t+1)}_{c} = \alpha \hat{S} \hat{\vy}^{(t)}_{c} + (1 - \alpha) \vy_c \quad \forall c \in \left\{1,...,C\right\}
    \label{equ:iterativelp}
\end{equation}
until convergence. Where $\alpha \in \left(0,1\right)$ is a propagation hyper-parameter, $\vy_c = \ve_c \in \left\{0,1\right\}^{N}$ is a one-hot vector with the non-zero element at index $c$, and $t$ is the current iteration. Prediction of the label for an unlabeled node $j \in \left\{C+1,...,N\right\}$ is then given by  
\begin{equation}
    \hat{y}_j = \argmax_{c} \hat{\vy}_c(j) ,
    \label{equ:predictionlp}
\end{equation}
where $\hat{\vy}_c(j) = \ve_j^{T} \hat{\vy}_c$ is the $j$-th element of the vector $\hat{\vy}_c$.
One can show~\cite{zbl+03} that this iterative solution is equivalent to solving $C$ linear systems
\begin{equation}
    L \hat{\vy}_c = \vy_c \quad \forall c \in \left\{1,...,C\right\} ,
    \label{equ:cglp}
\end{equation}
where $L = I - \alpha \hat{S}$ is the graph Laplacian. These linear systems have a closed-form solution
\begin{equation}
    \hat{\vy}_c = L^{-1} \vy_c  =  \Linv \vy_c .
    \label{equ:closedformlp}
\end{equation}
However, this closed-form solution is not practical for large datasets as the inverse graph Laplacian \Linv is a non-sparse $\real^{N \times N}$ matrix. For this reason it is usual~\cite{ita+19,gra+06,ita+16,chk+16} to solve~\equ{cglp} using the conjugate-gradient (CG) method, which is known to be faster than running the iterative solution~\cite{ita+16}. 
Using CG is possible because $L$ is positive-definite.

Observe that \equ{closedformlp} simply picks one of the columns of \Linv. Matrix element $\Linv(j, c)$ is the confidence of example $j$ belonging to class $c$. Its values are similarities, after label propagation, between each node pair. It is a type of geodesic similarity that captures the geometry of the feature space as this is indicated by the graph structure. Focusing on a classification task, we are only interested in similarities between an unlabeled example and a class node. 

\paragraph{Dual solution.} Herein, we show that solving $C$ linear systems of the form in~\equ{closedformlp} to obtain predictions for all unlabeled nodes using~\equ{predictionlp} is equivalent to solving $N-C$ linear systems of form
\begin{equation}
    \hat{\vz}_{j} = L^{-1} \ve_j \quad \forall j \in \left\{C+1,...,N\right\} ,
    \label{equ:duallp}
\end{equation}
and obtaining the unlabeled node prediction using
\begin{equation}
    \hat{y}_j = \argmax_{c}\hat{\vz}_j(c) .
    \label{equ:predictiondual}
\end{equation}
This comes from the fact that
\begin{equation}
    \hat{\vz}_j(c) = \ve_c^{T} L^{-1} \ve_j = \ve_j^{T} L^{-1} \ve_c = \hat{\vy}_c(j) .
    \label{equ:dualproof}
\end{equation}
Although we present the dual solution using the closed-form~\equ{closedformlp}, the same holds with the CG solution of~\equ{cglp}. Using the dual solution~\equ{duallp} is not practical for transductive learning as usually the unlabeled nodes are many more than the labeled ones. 
However, we show that this dual solution is efficiently used for inductive inference.

As discussed, we can view \Linv as a pairwise similarity matrix. The confidence of example $j$ belonging to class $c$, due to symmetry of \Linv, is equivalently obtained either by $\Linv(j,c)$ or $\Linv(c,j)$. This constitutes, an additional interpretation of the duality in the solution.

\paragraph{Inductive inference.} 
Test examples now come individually and are not known during  graph construction. 
A possible way to perform inductive inference is by adding the new node to the graph, which is expensive for a test-time operation as
$\hat{S}$ would have to be updated for each new test example.
Instead, inspired by~\cite{ita+16} that uses LP for retrieval,
we construct indicator vector $\vy_{\vx} \in \real^{N}$ 
for test example $\vx$
such that
\begin{equation}
    \vy_{\vx}(j) = \begin{cases}
        \vx^{T} \vx_j, & \text{if $\vx_j \in$ \knn($\vx$)} \\
        0, & \text{otherwise}.
    \end{cases} 
    \label{equ:inductivey}
\end{equation}
Then, we solve linear system 
\begin{equation}
\hat{\vz}_{\vx} = L^{-1} \vy_{\vx},
\label{equ:inductivez}
\end{equation}
as in the dual formulation~\equ{duallp} and get a prediction with $\hat{y}_{\vx} = \argmax_{c}\hat{\vz}_{\vx}(c)$. With the usual formulation of label propagation in~\equ{closedformlp}, $C$ linear systems need to be solved to get prediction for a single test example.
The dual formulation allows us to do it by solving only a single linear system.

\paragraph{Fast inductive inference with sparsification.} We further introduce an additional off-line step, where we solve~\equ{closedformlp}, get $\hat{\vy}_c$ for all $c \in \left\{1,...,C\right\}$, and store them in a matrix $\hat{Y} = \left[\hat{\vy}_1;...;\hat{\vy}_c\right] \in \real^{N \times C}$.
Then, the solution for a test example is equivalent to a weighted sum of rows of $\hat{Y}$~\cite{bdl06}, which is a byproduct of using the indicator vector~\equ{inductivey} for representing a test example.
Its prediction is given by $\hat{\vz}_{\vx} = \vy_{\vx}^{T} \hat{Y}$, and is equivalent to that obtained via \equ{inductivez}. However, storing the whole $\hat{Y}$ can be expensive for very large values of $N$ and $C$. 
We propose to sparsify $\hat{Y}$ by keeping only the largest values in each row, column, or over the whole matrix. 

Note that~\cite{iat+17} proposes a low-rank decomposition of the inverse graph Laplacian for the task of retrieval. 
Our solution is tailored to zero-shot classification, and we choose to obtain and sparsify the first $C$ rows of \Linv instead of approximating the whole matrix. Additionally, our solution requires one (sparse) vector to matrix multiplications at test-time instead of two.

\begin{figure*}[t]
     \centering
         \includegraphics[width=.42\textwidth]{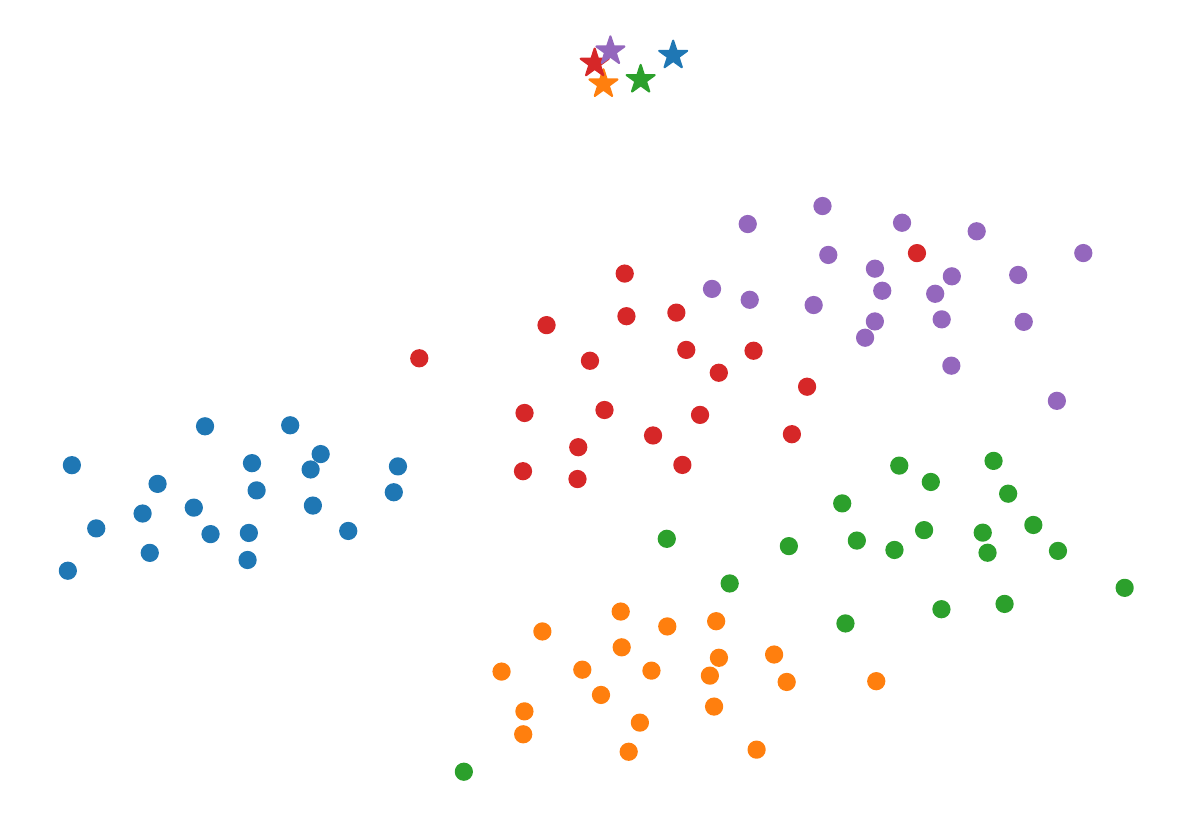}
         \hfill
         \includegraphics[width=.42\textwidth]{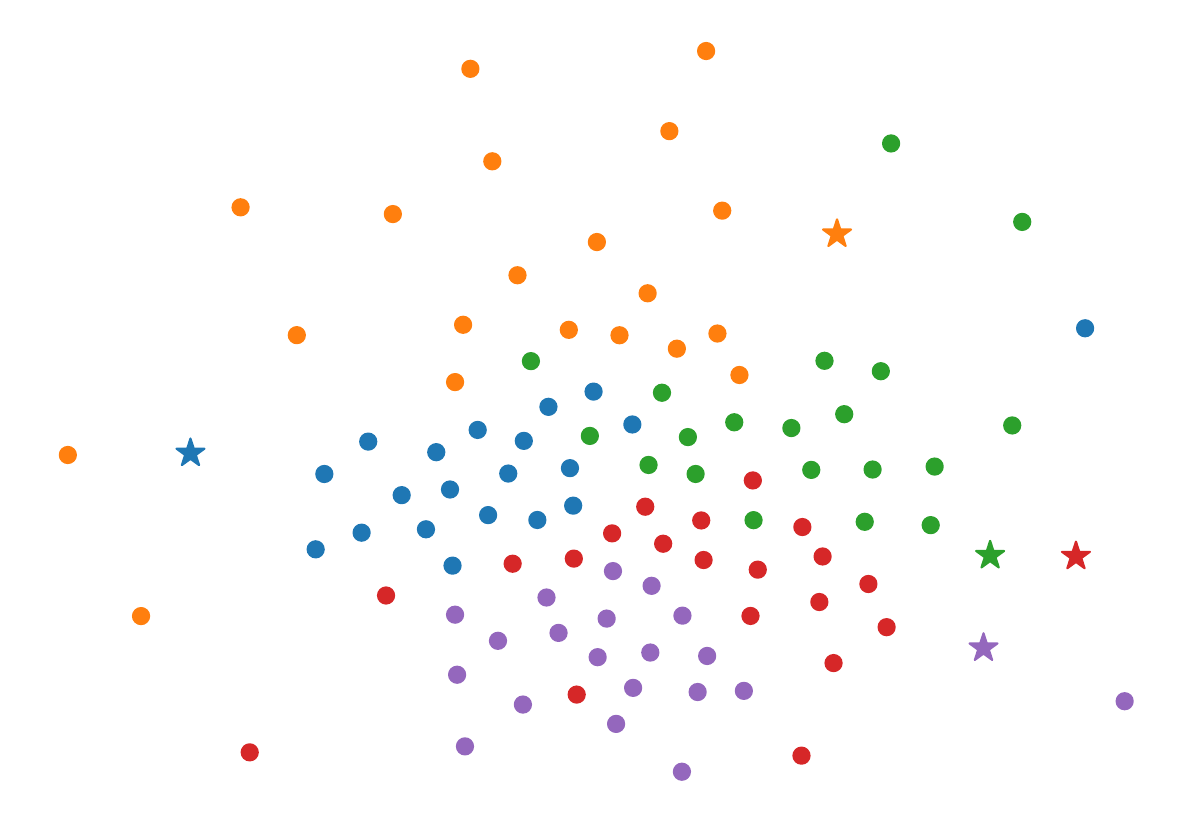}
    \caption{\textbf{t-SNE visualization} for the original CLIP  features (left) and our geodesic similarity (right). 
    The former is estimated with the features as input, while the latter with the \Linv used as a pairwise similarity matrix.    
    $\star$: class representation, $\bullet$: image representation. 
    Figure generated for five random classes from the CUB dataset.
    }
    \label{fig:tsne}
\end{figure*}

\subsection{LP for zero-shot VLM classification}

We are given a set of classes $\cC$ with extracted VLM representations $\left\{\vw_1,...,\vw_C\right\}$ and a set of unlabeled images $\cU$ with extracted VLM representations $\left\{\vu_1,...,\vu_M\right\}$.
We use them as nodes $\left\{\vw_1,...,\vw_C,\vu_1,...,\vu_M\right\}$ of the graph for label propagation. 
Nodes of class representations (text nodes) are labeled and image nodes unlabeled. \looseness=-1

\begin{figure}[t]
  \centering
  \input{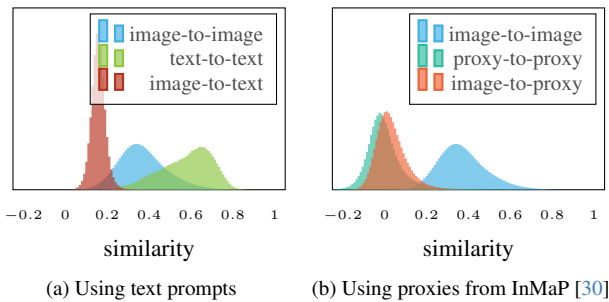}
\pgfplotsset{select coords between index/.style 2 args={
    x filter/.code={
        \ifnum\coordindex<#1\def\pgfmathresult{}\fi
        \ifnum\coordindex>#2\def\pgfmathresult{}\fi
    }
}}
\pgfplotsset{minor grid style={solid,gray,opacity=0.1}}
\pgfplotsset{major grid style={solid,gray,opacity=0.1}}
\pgfplotsset{every tick label/.append style={font=\tiny}}
\begin{subfigure}[b]{.48\linewidth}
 \centering
\begin{tikzpicture}
\begin{axis}[%
    width=1.3\linewidth,
    height=\linewidth,
    ybar,
    xmax = 1.05,
    xmin = -0.25,
    ymax = 15,
    xtick style={draw=none},
    ytick style={draw=none},
    ymin = 0,
    grid=none,
    yticklabels={,,},
    title style={yshift=-1.4ex},
    xlabel={\small{similarity}},%
    legend pos= {north east},
    legend style={cells={anchor=east}, font =\footnotesize, fill opacity=0.7, row sep=-2.5pt},
]
\addplot[color=CornflowerBlue, fill = CornflowerBlue,fill opacity=0.65,draw=none,bar width = 0.01] table[x expr={\thisrow{bins}}, y expr={\thisrow{im2im_hist} * 100}]  \simhist; \addlegendentry{image-to-image}
\addplot[color=LimeGreen, fill = LimeGreen,fill opacity=0.65,draw=none,bar width = 0.01] table[x expr={\thisrow{bins}}, y expr={\thisrow{text2text_hist} * 100}]  \simhist; \addlegendentry{text-to-text}
\addplot[color=BrickRed, fill = BrickRed,fill opacity=0.65,draw=none,bar width = 0.01] table[x expr={\thisrow{bins}}, y expr={\thisrow{im2text_hist} * 100}]  \simhist; \addlegendentry{image-to-text}
\end{axis}
\end{tikzpicture}

\caption{Using text prompts}
     \label{fig:similarities_clip}
\end{subfigure}
\hfill
\begin{subfigure}[b]{.48\linewidth}
 \centering
     
\begin{tikzpicture}
\begin{axis}[%
    width=1.3\textwidth,
    height=\linewidth,
    ybar,
    xmax = 1.05,
    xmin = -0.25,
    ymax = 15,
    xtick style={draw=none},
    ytick style={draw=none},
    ymin = 0,
    grid=none,
    title style={yshift=-1.4ex},    
    yticklabels={,,},
    xlabel={\small{similarity}},%
    legend pos= {north east},
    legend style={cells={anchor=east}, font =\footnotesize, fill opacity=0.7, row sep=-2.5pt},
]
\addplot[color=CornflowerBlue, fill = CornflowerBlue,fill opacity=0.65,draw=none,bar width = 0.01] table[x expr={\thisrow{bins}}, y expr={\thisrow{im2im_hist} * 100}]  \simhist; \addlegendentry{image-to-image}
\addplot[color=SeaGreen, fill = SeaGreen,fill opacity=0.65,draw=none,bar width = 0.01] table[x expr={\thisrow{bins}}, y expr={\thisrow{text2text_inmap_hist} * 100}]  \simhist; 
\addlegendentry{proxy-to-proxy}
\addplot[color=RedOrange, fill = RedOrange,fill opacity=0.65,draw=none,bar width = 0.01] table[x expr={\thisrow{bins}}, y expr={\thisrow{im2text_inmap_hist} * 100}]  \simhist; 
\addlegendentry{image-to-proxy}
\end{axis}
\end{tikzpicture}
\caption{Using proxies from InMaP~\cite{qxh23}}
     \label{fig:similarities_inmap}
 \end{subfigure}
  \caption{\textbf{Similarity distributions} among features of the same or different modality, using 7 textual templates~\cite{uga23} (left) or the InMaP proxies (right) as class representations.
  \label{fig:hist}}
\end{figure}

To construct the adjacency matrix $S$, we need to perform the $k$-nearest neighbor search between nodes. 
However, it is known that there exists a large modality gap between image and text representations coming from VLMs~\cite{lzk+22,uga23,zzh+23}. 
The respective similarity distributions for CLIP are shown in Figure~\ref{fig:similarities_clip}.
This modality gap makes standard kNN search between nodes not useful for label propagation; image nodes mostly get connected to image nodes, and text nodes mostly get connected to text nodes. 
As a consequence, few edges exist between labeled and unlabeled nodes.

To alleviate this problem, we perform the kNN search \textit{separately}
for connecting image nodes to image nodes and for connecting image nodes and text nodes. 
We do not perform the search using text nodes as queries, \ie text nodes get linked only if they appear in the kNN list of an image. 
This way we also avoid linking text nodes with each other, which is beneficial as each of them is labeled to a different class.
Formally, the values of the adjacency matrix are
\vspace{-1pt}
\begin{equation}
    s_{ij} = \begin{cases}
        \vu_i^{T} \vu_j, & \text{if $\vu_j \in \knn_{\vu} (\vu_i$)} \\
        \vu_i^{T} \vw_j,   & \text{if $\vw_j \in \knn_{\vw}(\vu_i$)} \\
        0, & \text{otherwise},
    \end{cases}
    \label{equ:adj_bi}
\end{equation}
\vspace{-1pt}
where $\knn_{\vu}$ and $\knn_{\vw}$ denote that the search is performed within the image or class features only, respectively.

Moreover, during inductive inference for image $\vu$, we perform the kNN search in a similar way to construct indicator vector $\vy_{\vu}$ whose elements are given by
\vspace{-1pt}
\begin{equation}
    \vy_{\vu}(i) = \begin{cases}
        \vu^{T} \vu_j, & \text{if $\vu_j \in \knn_{\vu}(\vu$)} \\
        \vu^{T} \vw_j, & \text{if $\vw_j \in \knn_{\vw}(\vu$)} \\
        0, & \text{otherwise}.
    \end{cases} 
    \label{equ:inductivey_bi}
\end{equation}
\vspace{-1pt}

Due to the two types of edges, \ie image-to-image and image-to-text, 
we use power function $h(v)=v^\gamma$ to transform the image-to-text (cross-modal) similarities. This way we effectively \textit{balance} their contribution in the graph and the indicator vector.
To that end, we use $h(\vu_i^{T} \vw_j)$ and $h(\vu^{T} \vw_j)$ in \equ{adj_bi} and \equ{inductivey_bi}, respectively, instead of $\vu_i^{T} \vw_j$ and $\vu^{T} \vw_j$.

We refer to the proposed method described above as \textbf{Z}ero-shot classification with \textbf{La}bel \textbf{P}ropagation (\textbf{\ours}). 
We further denote the variant of our method after sparsifying the $\hat{Y}$ matrix for inductive inference as \textbf{\ourssparse}.

\paragraph{$t$-SNE visualization of the bi-modal space.}
In Figure~\ref{fig:tsne} we visualize the bi-modal feature space for CLIP features  using $t$-SNE~\cite{tsne} in two cases, \ie the Euclidean case and using geodesic similarities obtained by \Linv, \ie after label propagation. When using  Euclidean affinities (left), we see that due to the large differences in the similarity distributions  (text-text, image-to-image, and text-to-image, as  shown in ~\Cref{fig:similarities_clip}) all class representations (stars) are clustered together  far from the image nodes. However, using the geodesic affinities from \Linv (right) we see that class representations are more spread.

\section{Experiments}
\label{sec:exp}

In this section, we first present the datasets we use, our experimental setup and competing methods. We then present component analysis for \ours and results for transductive and inductive zero-shot classification on 14 datasets.

\subsection{Datasets}

We evaluate the proposed method on 14 diverse image classification datasets: ImageNet ILSVRC2012~\cite{rds+15}, Describable Textures Dataset (DTD)~\cite{cmk+14}, EuroSAT~\cite{hbd+19}, FGVC-Aircraft~\cite{mrk+13}, Oxford Flowers 102~\cite{nz08}, Food-101~\cite{bgg14}, Oxford-IIIT Pet~\cite{pvz+12}, SUN397~\cite{xhe+10}, Stanford Cars~\cite{ksd+13}, Caltech101~\cite{ffp04}, UCF101~\cite{szs12}, CIFAR10~\cite{kri09}, CIFAR100~\cite{kri09}, CUB-200-2011~\cite{wbm+10}. For the first 11 datasets we borrow the train and test splits from CoOp\cite{zyl+22}.
We use the official training and test splits for %
CIFAR10, CIFAR100 and CUB-200-2011.

\subsection{Experimental setup}
In the transductive (inductive) inference setup, unlabeled nodes in the graph are the test (train) images. We always measure classification accuracy over the test images. 

\looseness=-1
\paragraph{VLMs and textual prompts.} 
We report results using the publicly available ResNet50 and ViT-B/16  CLIP~\cite{rkh+21} models. 
We adopt the 7 templates from SuS-X~\cite{uga23} as class prompts for all results apart from~\Cref{tab:cupl} where we utilize the LLM generated prompts from~\cite{plf22}.

\paragraph{Compared methods.} Our baseline is zero-shot recognition with \textbf{CLIP}~\cite{rkh+21} using text encoder features as class representations. \textbf{TPT}~\cite{snh+22} is based on test-time prompt tuning such that different image augmentations produce consistent predictions. 
The aforementioned methods do not exploit unlabeled data; their performance is therefore unchanged in both inference setups.
\textbf{CLIP-DN}~\cite{zrl+23} normalizes feature distributions during test-time and assumes access to the mean feature vector of the target distribution. In the transductive (inductive) setup the mean vector is estimated on the test (training) set. \textbf{InMaP}~\cite{qxh23} is a concurrent work that extracts updated class representations using pseudo-labels on the unlabeled set. In the transductive (inductive) setup the learning is performed on the test (training) images.

\paragraph{Implementation details.} 
We reproduce results for CLIP\footnote{\url{https://github.com/OpenAI/CLIP}}, CLIP-DN\footnote{\url{https://github.com/fengyuli-dev/distribution-normalization}}, and InMaP\footnote{\url{https://github.com/idstcv/InMaP}\label{foot:inmap}} using their public implementations. For TPT~\cite{snh+22} we report the numbers provided in~\cite{qxh23}. 
We run InMaP using a single set of hyper-parameters for all 14 datasets, \ie the default values reported in the official implementation\footref{foot:inmap}.
We also fix the values of $k$, $\gamma$, and $\alpha$ for \ours across all datasets to 5, 5.0, and 0.3, respectively, for CLIP, and 10, 3.0, and 0.3, respectively, for InMaP.

\paragraph{\ours variants.} We refer to \ours using text class representations as \textbf{CLIP + \ours}. Since InMaP is complementary to our work, we further evaluate the performance of \ours when $\{\vw_1, \ldots, \vw_C\}$ are the InMaP proxies. We refer to this as \textbf{InMaP + \ours} in the results. We refer to \ours with a sparse $\hat{Y}$ for inductive inference as \textbf{\ourssparse}.

\subsection{Components of \ours}

\paragraph{Bi-modal graph adjustments.} In Table~\ref{tab:search_abl} we show the importance of two design choices to adapt LP to bi-modal graphs, \ie separating the nearest neighbor search 
across
modalities using \equ{adj_bi} and \equ{inductivey_bi}, and transforming cross-modal similarities using a power function $h(\cdot)$. 
We see that separate search is crucial; without it LP is not effective at all. 
The power function gives an extra boost in both setups, especially in the case of transductive inference. 
In Table~\ref{tab:graph_edge} we report the percentage of images that are connected to their groundtruth class nodes within a path of length $n$, with and without our adjustments. We see that for any such paths to exist for the case without adjustments, $k$ needs to be extremely high. With adjustments, $k=5$ is enough for 71.4\% of the nodes to be connected to the correct class nodes.

\paragraph{Sparsifying matrix $\hat{Y}$ for inductive inference.}

\begin{figure*}[t]
  \centering
  \input{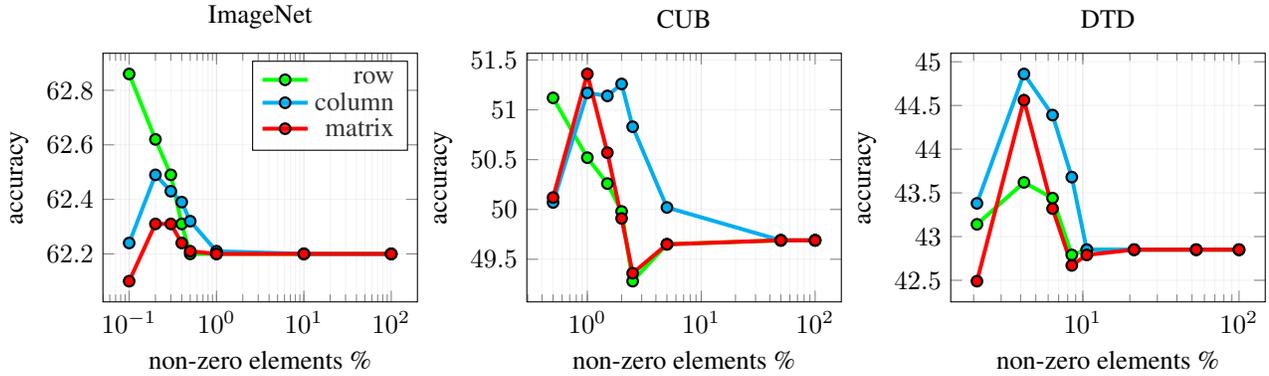}
\pgfplotsset{every tick label/.append style={font=\normalsize}}
\pgfplotsset{select coords between index/.style 2 args={
    x filter/.code={
        \ifnum\coordindex<#1\def\pgfmathresult{}\fi
        \ifnum\coordindex>#2\def\pgfmathresult{}\fi
    }
}}
\pgfplotsset{minor grid style={solid,gray,opacity=0.1}}
\pgfplotsset{major grid style={solid,gray,opacity=0.1}}
\begin{tikzpicture}
\begin{axis}[%
	width=.33\linewidth,
	height=.28\linewidth,
	xlabel={non-zero elements \%},
	ylabel={accuracy},
    xmode=log,
    legend pos= {north east},
    legend style={cells={anchor=east}, font =\normalsize, fill opacity=0.8, row sep=-2.5pt},
    title style={align=center, row sep=10pt},
    title={{ImageNet}},
    grid=both,
]
	\addplot[mark options={draw=black, line width=0.8},color=green,     solid, mark=*,   mark size=2.0, line width=1.5, select coords between index={0}{10}] table[x=percentage, y=acc_cls] \approximationimagenet;\addlegendentry{row};
    \addplot[mark options={draw=black, line width=0.8},color=cyan,     solid, mark=*,   mark size=2.0, line width=1.5, select coords between index={0}{10}] table[x=percentage, y=acc_img] \approximationimagenet;\addlegendentry{column};
    \addplot[mark options={draw=black, line width=0.8},color=red,     solid, mark=*,   mark size=2.0, line width=1.5, select coords between index={0}{10}] table[x=percentage, y=acc_all] \approximationimagenet;\addlegendentry{matrix};
\end{axis}
\end{tikzpicture}
\begin{tikzpicture}
\begin{axis}[%
	width=.33\linewidth,
	height=.28\linewidth,
	xlabel={non-zero elements \%},
	ylabel={accuracy},
    xmode=log,
    legend pos= {north east},
    legend style={cells={anchor=east}, font =\normalsize, fill opacity=0.8, row sep=-2.5pt},
    title style={align=center, row sep=10pt},
    title={{CUB}},
    grid=both,
]
	\addplot[mark options={draw=black, line width=0.8},color=green,     solid, mark=*,   mark size=2.0, line width=1.5, select coords between index={0}{10}] table[x=percentage, y=acc_cls] \approximationcub;%
    \addplot[mark options={draw=black, line width=0.8},color=cyan,     solid, mark=*,   mark size=2.0, line width=1.5, select coords between index={0}{10}] table[x=percentage, y=acc_img] \approximationcub;%
    \addplot[mark options={draw=black, line width=0.8},color=red,     solid, mark=*,   mark size=2.0, line width=1.5, select coords between index={0}{10}] table[x=percentage, y=acc_all] \approximationcub;%
\end{axis}
\end{tikzpicture}
\begin{tikzpicture}
\begin{axis}[%
	width=.33\linewidth,
	height=.28\linewidth,
	xlabel={non-zero elements \%},
	ylabel={accuracy},
    xmode=log,
    legend pos= {north east},
    legend style={cells={anchor=east}, font =\normalsize, fill opacity=0.8, row sep=-2.5pt},
    title style={align=center, row sep=10pt},
    title={{DTD}},
    grid=both,
]
	\addplot[mark options={draw=black, line width=0.8},color=green,     solid, mark=*,   mark size=2.0, line width=1.5, select coords between index={0}{10}] table[x=percentage, y=acc_cls] \approximationdtd;%
    \addplot[mark options={draw=black, line width=0.8},color=cyan,     solid, mark=*,   mark size=2.0, line width=1.5, select coords between index={0}{10}] table[x=percentage, y=acc_img] \approximationdtd;%
    \addplot[mark options={draw=black, line width=0.8},color=red,     solid, mark=*,   mark size=2.0, line width=1.5, select coords between index={0}{10}] table[x=percentage, y=acc_all] \approximationdtd;%
\end{axis}
\end{tikzpicture}
  \caption{\textbf{Sparcifying matrix $\hat{Y}$ for inductive CLIP+\ours:} effect of maintaining only the top elements per row/column/matrix. \label{fig:approx}}
\end{figure*}

\begin{figure*}[t]
\input{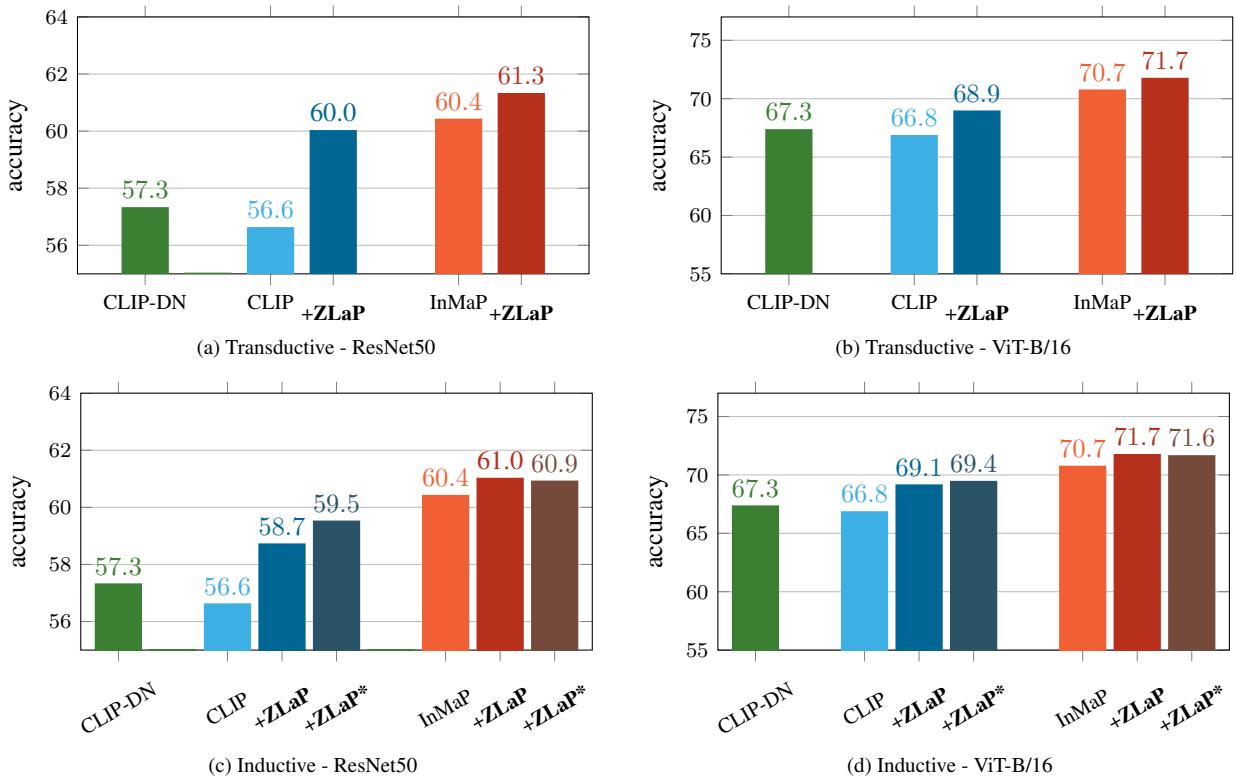}
    \caption{\textbf{Zero-shot classification accuracy averaged over 14 datasets} for the transductive (top) and inductive (bottom) setups.
    Results per dataset are reported in the supplementary material. 
    \label{fig:results} }
\end{figure*}

We explore three ways of approximating $\hat{Y}$ by sparsification, \ie wither keeping only the largest $\xi$ columns per row, the largest $\xi$ rows per column, or the largest $\xi$ elements of the whole matrix. In all cases, the rest of the elements are set to zero. In Figure~\ref{fig:approx} we show the influence that these three variants have on performance. %
Not only these variants speed-up inference, but we can also see improvements in performance when sparsification percentage is high, \ie low percentage of non-zero elements.
We attribute this to the fact that less confident predictions in $\hat{Y}$, many of them erroneous, are now set to zero.
Although the best variant to choose seems to vary per dataset, we found that keeping the top element per row performs well across different datasets. 
We therefore use the $\xi=1$ top element per row for our experiments. This amounts to different percentages of sparsity per dataset; we are keeping approximately $2.3\%$ on average across all datasets.
Regarding the inference speed-up, the primal solution takes $\sim 2.6$ sec per image, the dual takes $\sim 4.4$ ms, while the sparsified approach takes $\sim 0.6$ ms, measured on ImageNet dataset.

\begin{table}
    \begin{subtable}{.47\textwidth}
        \centering
        \newcommand{\cmark}{\ding{51}}%
\newcommand{\xmark}{\ding{55}}

\small
\begin{tabular}{Wc{1.2cm}Wc{1.2cm}Wr{1.2cm}Wr{1.2cm}Wr{1.2cm}}
\toprule
Eq.\equ{adj_bi} & $h(\cdot)$ & ImageNet & DTD & CUB \\
\midrule
\xmark               & \xmark           & 0.1                          & 2.1                    & 0.5                    \\
\xmark               & \cmark           & 0.1                          & 2.1                    & 0.5                    \\
\cmark               & \xmark           & 50.2                        & 32.3                   & 41.6                   \\
\cmark               & \cmark           & 61.8                        & 41.9                   & 52.1      \\
\bottomrule
\end{tabular}

        \caption{Transductive inference}
    \end{subtable}
    \newline
    \vspace{2pt}
    \newline
    \begin{subtable}{.47\textwidth}
        \centering
        \newcommand{\cmark}{\ding{51}}%
\newcommand{\xmark}{\ding{55}}

\small
\begin{tabular}{Wc{1.2cm}Wc{1.2cm}Wr{1.2cm}Wr{1.2cm}Wr{1.2cm}}
\toprule
Eq.\equ{adj_bi}-\equ{inductivey_bi} & $h(\cdot)$ & ImageNet & DTD & CUB \\
\midrule
\xmark               & \xmark           & 0.1                          & 2.1                    & 0.5                    \\
\xmark               & \cmark           & 0.1                          & 2.1                    & 0.5                    \\
\cmark               & \xmark           & 60.8                        & 42.4                   & 49.6                   \\
\cmark               & \cmark           & 62.2                        & 42.8                   & 49.7      \\
\bottomrule
\end{tabular}

        \caption{Inductive inference}
    \end{subtable}
    \caption{\textbf{Adjusting LP to bi-modal graphs.} Impact of using separate kNN search for constructing the graph (Eq.\equ{adj_bi}) or the indicator vector (Eq.\equ{inductivey_bi}), as well as power function $h(\cdot)$ for balancing the contributions of two types of edges in the graph.}
    \label{tab:search_abl}
\end{table}

\begin{table}
    \centering
    \small
\begin{tabular}{lrrr@{\xlsp}rrr}
\toprule
 & \multicolumn{3}{c}{Joint kNN search}                            & \multicolumn{3}{c}{Separate kNN search}                         \\
 
\cmidrule{2-7}
\cmidrule{2-7}
$k$                  & $n$=1 & $n$=2 & $n$=3 & $n$=1 & $n$=2 & $n$=3      \\

\cmidrule{2-7}
5                  & 0.0        & 0.0       & 0.0      & 71.4     & 85.1    & 100.0    \\
10                 & 0.0       & 0.0       & 0.0      & 82.9     & 95.4    & 100.0    \\
100                & 40.1     & 100.0     & 100.0    & 100.0      & 100.0     & 100.0   \\
\bottomrule
\end{tabular}

    \caption{\textbf{Impact of the separate kNN search} on the shortest paths between image nodes and the text node of their class. 
    We report the percentage of images whose shortest path to the text node of their ground-truth class has length equal to or less than $n$.
    \textit{Left:} the vanilla approach. \textit{Right:} our separate kNN search using Eq.~\equ{adj_bi}.    
    Analysis on DTD for the transductive setup.}
    \label{tab:graph_edge}
\end{table}

\paragraph{Using the class proxies from InMaP.} We observe that many class-to-image similarities (\eg $\vu_i^\top \vw_j$ in \equ{adj_bi}) become negative on some datasets when using \ours with InMaP proxies (see~\Cref{fig:similarities_inmap}). We therefore perform min-max normalization in range $[0, 1]$ after constructing adjacency matrix $S$ or the indicator vector, for the transductive and inductive inference setups.

\subsection{Results}

\paragraph{Transductive inference.} We present results for transductive zero-shot classification in Figure~\ref{fig:results}. \ours improves the zero-shot performance of CLIP significantly on all datasets. It also outperforms the recent TPT and CLIP-DN approaches on the vast majority of cases, with large gains in average accuracy. Compared to InMaP, \ours offers lower accuracy on average. However, by incorporating InMaP's class representations to our graph, we can improve our results even further and outperform all other methods, for an improvement of approximately +5\% over CLIP with both backbones.
\looseness=-1

\paragraph{Inductive inference.} We report results for the inductive inference setup in Figure~\ref{fig:results}. 
\ours achieves a noticeable improvements over the CLIP baseline in this setup as well. Gains are more prominent for the case of \ours using the InMaP proxies, where gains over CLIP are  +4.4\% and +4.9\% for the two backbones. We also observe that, although InMaP slightly outperforms \ours when used over CLIP, the combination of the two achieves \textit{state-of-the-art performance} in this case as well.
We further see that \ourssparse retains the state-of-the-art performance of our method, while sparsifying $\hat{Y}$ offers significant speed-up at inference time.

\begin{table} [t]
    \centering
    \small
\setlength{\tabcolsep}{20pt} %
\renewcommand{\arraystretch}{1.1} %
\begin{tabular}{lcc}
\toprule
 & Transductive & Inductive \\
\midrule
\multicolumn{3}{l}{\textit{Results with RN50}} \\
~CLIP               &  63.0 & 63.0                    \\
\hspace{0.5em} + \textbf{\ours}               &  64.6 &  64.2  \\
~InMaP              &   \underline{64.8}  &  \underline{64.6}    \\
\hspace{0.5em} + \textbf{\ours}               &  \textbf{65.8} &  \textbf{65.0}  \\
\midrule
\multicolumn{3}{l}{\textit{Results with ViT-B-16}} \\
~CLIP               &  71.9 & 71.9                    \\
\hspace{0.5em} + \textbf{\ours}               &  72.6 &  73.3  \\
~InMaP              &   \underline{73.9}  &  \underline{74.0}    \\
\hspace{0.5em} + \textbf{\ours}               &  \textbf{74.8} &  \textbf{74.2}  \\
\bottomrule
\end{tabular}
     \caption{\textbf{Zero-shot classification using prompts generated by LLMs~\cite{plf22}.} We report average accuracy on 12 datasets using prompts from CuPL~\cite{plf22} together with our 7 standard prompts. 
     Results per dataset are reported in the supplementary material.
     }
    \label{tab:cupl}    
\end{table}

\paragraph{Leveraging LLM generated prompts.} In Table~\ref{tab:cupl} we report average zero-shot classification accuracy for \ours using the prompts recently proposed in CuPL~\cite{plf22}. These are prompts generated by LMMs that are available on the CuPL Github page\footnote{\url{https://github.com/sarahpratt/CuPL}} for 12 of the datasets we use (all datasets besides CUB and Eurosat). \ours improves zero-shot performance in this case as well, for both the transductive and inductive setups. This verifies that our method is complementary to improved prompt engineering. 
\looseness=-1

\paragraph{Multi-label classification.} We apply \ours for multi-label classification on the MS-COCO~\cite{lmb+14} dataset. \ours improves the zero-shot performance of CLIP by +6.0\% mAP (56.8\% vs. 50.8\%) for inductive inference without any modification of the approach or its hyper-parameters.
\looseness=-1

\paragraph{Web-crawled unlabeled images.} All previous experiments use unlabeled images that come from the target distribution, \ie they are known to depict one of the classes of interest but their labels are discarded. To see the impact of \ours in a more realistic setup using web-crawled images we rely on LAION-400M~\cite{sbv+22} composed by image-caption pairs. We construct the set of unlabeled images with 10,000 images per class that are chosen either randomly, or based on proximity of their image or text features to the class representation. Random selection fails, but the other two options provide some improvement compared to CLIP, with the caption-based neighbors being a bit better. The complete set of results is presented in the supplementary material.
\looseness=-1

\paragraph{Different VLMs}
We use CLIP as the VLM of choice throughout our experiments. In Table~\ref{tab:vlm}, we present  results when \ours is applied on top of four recent VLMs, namely BLIP~\cite{llx+22}, ALBEF~\cite{lsg+21}, and two versions of EVA-CLIP~\cite{swy+24}. We use the implementations of BLIP and ALBEF that are available in the LAVIS library\footnote{\url{https://github.com/salesforce/LAVIS}}, while for EVA-CLIP we use implementation from the official Github repository\footnote{\url{https://github.com/baaivision/EVA/tree/master/EVA-CLIP-18B}}. \ours improves the results of all four different VLMs in both transductive and inductive setups.

\begin{table}[t]
    \centering
    \small
\begin{tabular}{lcc}
\toprule
 & Transductive & Inductive \\
\midrule
~BLIP~\cite{llx+22}               &  54.6 & 54.6                    \\
\hspace{0.5em} + \textbf{\ours}               &  \textbf{59.6} &  \textbf{57.9}  \\
\midrule
~ALBEF~\cite{lsg+21}               &  36.0 & 36.0                    \\
\hspace{0.5em} + \textbf{\ours}               &  \textbf{41.2} &  \textbf{46.8}  \\
\midrule
~EVA-CLIP-8B~\cite{swy+24}               &  83.6 & 83.6                    \\
\hspace{0.5em} + \textbf{\ours}               &  \textbf{84.6} &  \textbf{84.5}  \\
\midrule
~EVA-CLIP-18B~\cite{swy+24}               &  83.9 & 83.9                    \\
\hspace{0.5em} + \textbf{\ours}               &  \textbf{84.8} &  \textbf{84.7}  \\
\bottomrule
\end{tabular}
    \caption{\textbf{Accuracy on ImageNet using different VLMs.}}
    \label{tab:vlm}
\end{table}

\section{Conclusions}
\label{sec:conclusions}
Label propagation is an intuitive way of encoding the global structure of unlabeled data into geodesic distances over a locally Euclidean space. In this paper, we show that this method can be successfully tailored to both transductive and inductive zero-shot classification with vision-language models, and achieve state-of-the-art performance on both setups. To that end, we show that it is highly important to take proper care of the peculiarities of the bi-modal nature of the task during graph construction. We further carefully design an efficient variant of label propagation for the inductive inference case, that may enable label propagation to be applied to other tasks 
beyond zero-shot classification.

Vision-language models trained on billion-scale datasets are redefining computer vision research. The proposed \ours is a training-free approach able to improve the generalization performance of black-box VLMs using only unlabeled data, for an annotation-free, text-based and open-world classification paradigm that will inevitably be ubiquitous in the near future.

\paragraph{Acknowledgements.} This work was supported by the Junior Star GACR GM 21- 28830M and the Czech Technical University in Prague grant No. SGS23/173/OHK3/3T/13. We thank Ahmet Iscen for many helpful comments.

\clearpage
{\small
\bibliographystyle{ieeenat_fullname}
\bibliography{tex/bib}
}

\clearpage
\newpage
\section*{Appendix}
\appendix
\paragraph{Impact of hyper-parameters}

We show the effect of $\gamma$ and $\alpha$ in Table~\ref{tab:gamma_alpha} and observe stability over a wide range of values for both cases.

\begin{table}[b]
    \centering
    \footnotesize
\begin{tabular}{rrrrrrrrr}
\hline
\multicolumn{1}{l}{\diagbox[innerwidth=0.4cm]{$\gamma$}{$\alpha$}} & 0.01                         & 0.05                         & 0.1                          & 0.3                          & 0.5                          & 0.7                          & 0.8                          & 0.9                          \\
\hline
0.5                  & \cellcolor[HTML]{F6D1CE}33.6 & \cellcolor[HTML]{F6D3D1}34.4 & \cellcolor[HTML]{F7D6D3}35.2 & \cellcolor[HTML]{F8DDDA}37.9 & \cellcolor[HTML]{F9E2E0}39.9 & \cellcolor[HTML]{FAE9E8}42.8 & \cellcolor[HTML]{FBECEB}43.9 & \cellcolor[HTML]{FBEBEA}43.6 \\
1                    & \cellcolor[HTML]{F8DBD9}37.5 & \cellcolor[HTML]{F8DEDB}38.3 & \cellcolor[HTML]{F9E0DE}39.2 & \cellcolor[HTML]{FAE6E5}41.6 & \cellcolor[HTML]{FBECEB}43.8 & \cellcolor[HTML]{FCF2F1}46.0 & \cellcolor[HTML]{FCF4F3}46.8 & \cellcolor[HTML]{FCF1F0}45.9 \\
5                    & \cellcolor[HTML]{FEFEFE}50.9 & \cellcolor[HTML]{EEF8F3}51.3 & \cellcolor[HTML]{E0F3E9}51.5 & \cellcolor[HTML]{BBE4D0}52.1 & \cellcolor[HTML]{A6DBC1}52.5 & \cellcolor[HTML]{B4E1CB}52.3 & \cellcolor[HTML]{E4F4EC}51.4 & \cellcolor[HTML]{FEFBFB}49.7 \\
8                    & \cellcolor[HTML]{B9E3CE}52.2 & \cellcolor[HTML]{9AD6B9}52.7 & \cellcolor[HTML]{8FD2B1}52.9 & \cellcolor[HTML]{5DBE8E}53.7 & \cellcolor[HTML]{65C194}53.6 & \cellcolor[HTML]{A1D9BE}52.6 & \cellcolor[HTML]{E1F3EA}51.5 & \cellcolor[HTML]{FEFBFB}49.6 \\
10                   & \cellcolor[HTML]{95D4B5}52.8 & \cellcolor[HTML]{90D2B2}52.9 & \cellcolor[HTML]{7ECBA6}53.2 & \cellcolor[HTML]{57BB8A}53.8 & \cellcolor[HTML]{64C093}53.6 & \cellcolor[HTML]{AADDC4}52.4 & \cellcolor[HTML]{E4F4EC}51.4 & \cellcolor[HTML]{FEFAF9}49.2 \\
12                   & \cellcolor[HTML]{94D4B5}52.8 & \cellcolor[HTML]{96D5B6}52.8 & \cellcolor[HTML]{84CDA9}53.1 & \cellcolor[HTML]{64C093}53.6 & \cellcolor[HTML]{76C8A0}53.3 & \cellcolor[HTML]{AEDEC6}52.4 & \cellcolor[HTML]{FCFEFD}51.0 & \cellcolor[HTML]{FEF9F9}49.0 \\
15                   & \cellcolor[HTML]{E67E75}1.5  & \cellcolor[HTML]{C1E6D4}52.0 & \cellcolor[HTML]{C2E7D5}52.0 & \cellcolor[HTML]{86CEAB}53.0 & \cellcolor[HTML]{93D4B4}52.8 & \cellcolor[HTML]{D4EEE2}51.7 & \cellcolor[HTML]{FEFDFD}50.4 & \cellcolor[HTML]{FDF8F8}48.6 \\
20                   & \cellcolor[HTML]{E67C73}0.5  & \cellcolor[HTML]{E67C73}0.5  & \cellcolor[HTML]{E67C73}0.5  & \cellcolor[HTML]{F3C0BC}27.0 & \cellcolor[HTML]{FDF7F7}48.1 & \cellcolor[HTML]{FDF5F5}47.5 & \cellcolor[HTML]{FCF2F1}46.2 & \cellcolor[HTML]{FBEFEE}44.9 \\
\hline
\end{tabular}
    \caption{\textbf{Impact of $\gamma$ and $\alpha$ hyper-parameters.} Results presented on CUB dataset for transductive inference.}
    \label{tab:gamma_alpha}
\end{table}
\paragraph{Additional backbones}

In Table~\ref{tab:backbones}, we present transductive and inductive zero-shot classification results on ImageNet with additional CLIP backbones. We present results with two versions of ViT-L-14 from CLIP~\cite{rkh+21}. Additionally, we present the results with ViT-B-16 and ViT-H-14 from OpenCLIP~\cite{cbw+23}\footnote{\url{https://github.com/mlfoundations/open_clip}} trained on the LAION-2B~\cite{sbv+22} dataset. We see from Table~\ref{tab:backbones} that \ours improves the results with different backbones in both transductive and inductive setups. This verifies that \ours is not backbone dependant and that it is independent of the dataset used for pre-training.

\begin{table}[t]
    \centering
    \small
\begin{tabular}{lcc}
\toprule
 & Transductive & Inductive \\
\midrule
\multicolumn{3}{l}{\textit{Results with ViT-L-14}} \\
~CLIP               &  75.9 & 75.9                    \\
\hspace{0.5em} + \textbf{\ours}               &  \textbf{77.2} &  \textbf{77.3}  \\
\midrule
\multicolumn{3}{l}{\textit{Results with ViT-L-14@336}} \\
~CLIP               &  77.0 & 77.0                    \\
\hspace{0.5em} + \textbf{\ours}               &  \textbf{78.0} &  \textbf{78.4}  \\
\midrule
\multicolumn{3}{l}{\textit{Results with ViT-B-16 (LAION-2B)}} \\
~CLIP               &  70.4 & 70.4                    \\
\hspace{0.5em} + \textbf{\ours}               &  \textbf{72.0} &  \textbf{72.1}  \\
\midrule
\multicolumn{3}{l}{\textit{Results with ViT-H-14 (LAION-2B)}} \\
~CLIP               &  78.0 & 78.0                    \\
\hspace{0.5em} + \textbf{\ours}               &  \textbf{79.1} &  \textbf{79.1}  \\
\bottomrule
\end{tabular}
    \caption{\textbf{Accuracy on ImageNet using different CLIP backbones.}}
    \label{tab:backbones}
\end{table}
\paragraph{Per-dataset results}

In Tables~\ref{tab:transductive} and~\ref{tab:inductive}, we present per dataset results for transductive and inductive setups, respectively.

\begin{table*}[t]
    \centering
    {
\footnotesize
\renewcommand*{\arraystretch}{1.2}
\begin{tabular}{lp{.58cm}p{.58cm}p{.58cm}p{.58cm}p{.58cm}p{.58cm}p{.58cm}p{.58cm}p{.58cm}p{.58cm}p{.58cm}p{.58cm}p{.58cm}p{.58cm}|p{.58cm}}
\toprule
        & \scriptsize  \rotatebox{30}{imagenet} & \scriptsize  \rotatebox{30}{dtd} & \scriptsize  \rotatebox{30}{eurosat} & \scriptsize  \rotatebox{30}{fgvca} & \scriptsize  \rotatebox{30}{flowers} & \scriptsize  \rotatebox{30}{food} & \scriptsize  \rotatebox{30}{pets} & \scriptsize  \rotatebox{30}{sun} & \scriptsize  \rotatebox{30}{cars} & \scriptsize  \rotatebox{30}{caltech} & \scriptsize  \rotatebox{30}{cifar10} & \scriptsize  \rotatebox{30}{cifar100} & \scriptsize  \rotatebox{30}{cub} & \scriptsize  \rotatebox{30}{ucf} & \scriptsize  \rotatebox{30}{avg} \\
\midrule
\multicolumn{16}{l}{\textit{Results with ResNet50}}                                                                                                             \\
~TPT\textsuperscript{\dag}    & 60.7                        & 40.8                   & 28.3                       & 17.6                     & 62.7                          & 74.9                       & 84.5                   & 61.5                      & 58.5                    & 87.0                          &              --               &        --                      &     --                    & \textbf{69.8}                      & \textit{(58.8)}                  \\
~CLIP-DN & 60.2 & 41.1 & 28.4 & 17.3 & 63.3 & 77.2 & 83.1 & 60.9 & 54.8 & \underline{88.3} & 74.0 & 44.7 & 48.9 & 60.4 & 57.3                   \\ 
~CLIP    & 60.3                        & 41.1                   & 26.9                       & 16.7                     & 62.9                          & 76.6                     & 83.1                   & 61.2                     & 54.4                    & 87.9                          & 72.3                       & 42.5                        & 47.0                   & 59.9                      & 56.6                   \\
\hspace{0.5em} + \textbf{\ours} & \underline{61.8} & 41.9 & \textbf{35.5} & 17.8 & \underline{65.9} & 78.8 & 83.9 & 63.3 & 57.8 & \textbf{89.6} & 78.2 & 47.9 & 52.1 & 65.9 & 60.0 \\ 
\quad \emph{(vs CLIP)} & \gain{1.5} & \gain{0.8} & \gain{8.6} & \gain{1.1} & \gain{3.0} & \gain{2.2} & \gain{0.8} & \gain{2.1} & \gain{3.4} & \gain{1.7} & \gain{5.9} & \gain{5.4} & \gain{5.1} & \gain{6.0} & \gain{3.4} \\
~InMaP   & \textbf{63.8} & \underline{44.8} & 33.4 & \textbf{19.0} & 65.0 & \textbf{79.4} & \underline{89.0} & \underline{65.3} & \underline{61.5} & 74.5 & \underline{78.9} & \underline{49.6} & \textbf{55.5} & 65.6 & \underline{60.4}                   \\ 
\hspace{0.5em} + \textbf{\ours} & \textbf{63.8} & \textbf{45.9} & \underline{34.5} & \underline{18.4} & \textbf{67.1} & \underline{79.2} & \textbf{89.2} & \textbf{65.9} & \textbf{62.0} & 80.7 & \textbf{79.2} & \textbf{49.7} & \underline{55.3} & \underline{67.8} & \textbf{61.3} \\
\quad \emph{(vs CLIP)} & \gain{3.5} & \gain{4.8} & \gain{7.6} & \gain{1.7} & \gain{4.2} & \gain{2.6} & \gain{6.1} & \gain{4.7} & \gain{7.6} & \loss{7.2} & \gain{6.9} & \gain{7.2} & \gain{8.3} & \gain{7.9} & \gain{4.7} \\
\quad \emph{(vs InMaP)} & \gain{0.0} & \gain{1.1} & \gain{1.1} & \loss{0.6} & \gain{2.1} & \loss{0.2} & \gain{0.2} & \gain{0.6} & \gain{0.5} & \gain{6.2} & \gain{0.3} & \gain{0.1} & \loss{0.2} & \gain{2.2} & \gain{0.9} \\
\midrule
\multicolumn{16}{l}{\textit{Results with ViT-B/16}} \\
~TPT\textsuperscript{\dag}    & 69.0 & 47.8 & 42.4 & 24.8 & 69.0 & 84.7 & 87.8 & 65.5 & 66.9 & \textbf{94.2} & --& --& --& 68.0 & \textit{(65.5)}                   \\
~CLIP-DN & 68.3 & 45.7 & 53.3 & 24.3 & 68.0 & 86.0 & 87.7 & 66.5 & 64.0 & 93.6 & 91.4 & 69.6 & 56.1 & 68.4 & 67.3                   \\ 
~CLIP    & 68.8 & 45.1 & 50.2 & 23.0 & 67.0 & 85.7 & 88.3 & 66.3 & 63.8 & \underline{93.9} & 91.2 & 68.7 & 55.2 & 67.5 & 66.8                   \\
\hspace{0.5em} + \textbf{\ours} & 69.7 & 46.0 & 57.7 & 26.3 & 67.9 & 87.2 & 87.9 & 67.8 & 66.8 & 91.8 & 92.6 & \underline{70.8} & 58.2 & 73.8 & 68.9 \\ 
\quad \emph{(vs CLIP)} & \gain{0.9} & \gain{0.9} & \gain{7.5} & \gain{3.3} & \gain{0.9} & \gain{1.5} & \loss{0.4} & \gain{1.5} & \gain{3.0} & \loss{2.1} & \gain{1.4} & \gain{2.1} & \gain{3.0} & \gain{6.3} & \gain{2.1} \\
~InMaP   & \underline{72.5} & \underline{50.9} & \underline{60.1} & \underline{28.3} & \underline{70.8} & \textbf{88.0} & \textbf{93.2} & \underline{71.3} & \underline{71.7} & 76.7 & \underline{93.3} & \textbf{73.3} & \underline{63.8} & \underline{75.7} & \underline{70.7}          \\ 
\hspace{0.5em} + \textbf{\ours} & \textbf{72.7} & \textbf{51.8} & \textbf{60.9} & \textbf{28.4} & \textbf{73.4} & \underline{87.9} & \underline{92.8} & \textbf{71.9} & \textbf{72.1} & 83.7 & \textbf{93.6} & \textbf{73.3} & \textbf{64.1} & \textbf{77.7} & \textbf{71.7} \\
\quad \emph{(vs CLIP)} & \gain{3.9} & \gain{6.7} & \gain{10.7} & \gain{5.4} & \gain{6.4} & \gain{2.2} & \gain{4.5} & \gain{5.6} & \gain{8.3} & \loss{10.2} & \gain{2.4} & \gain{4.6} & \gain{8.9} & \gain{10.2} & \gain{4.9}\\
\quad \emph{(vs InMaP)} & \gain{0.2} & \gain{0.9} & \gain{0.8} & \gain{0.1} & \gain{2.6} & \loss{0.1} & \loss{0.4} & \gain{0.6} & \gain{0.4} & \gain{7.0} & \gain{0.3} & \gain{0.0} & \gain{0.3} & \gain{2.0} & \gain{1.0}\\
\bottomrule
\end{tabular}
}

    \caption{\textbf{Trasductive zero-shot classification accuracy on 14 datasets} for two CLIP backbones.
    Rows denoted as \emph{(vs CLIP)} and \emph{(vs InMaP)} show the absolute accuracy gains of our method over CLIP and InMaP, respectively.
    \textsuperscript{\dag} denotes numbers taken from InMaP~\cite{qxh23}. 
    }
    \label{tab:transductive}
\end{table*}

\begin{table*}[b]
    \centering
    {
\renewcommand*{\arraystretch}{1.2}
\footnotesize
\begin{tabular}{lp{.58cm}p{.58cm}p{.58cm}p{.58cm}p{.58cm}p{.58cm}p{.58cm}p{.58cm}p{.58cm}p{.58cm}p{.58cm}p{.58cm}p{.58cm}p{.58cm}|p{.58cm}}
\toprule
        & \scriptsize  \rotatebox{30}{imagenet} & \scriptsize  \rotatebox{30}{dtd} & \scriptsize  \rotatebox{30}{eurosat} & \scriptsize  \rotatebox{30}{fgvca} & \scriptsize  \rotatebox{30}{flowers} & \scriptsize  \rotatebox{30}{food} & \scriptsize  \rotatebox{30}{pets} & \scriptsize  \rotatebox{30}{sun} & \scriptsize  \rotatebox{30}{cars} & \scriptsize  \rotatebox{30}{caltech} & \scriptsize  \rotatebox{30}{cifar10} & \scriptsize  \rotatebox{30}{cifar100} & \scriptsize  \rotatebox{30}{cub} & \scriptsize  \rotatebox{30}{ucf} & \scriptsize  \rotatebox{30}{avg} \\
\midrule
\multicolumn{16}{l}{\textit{Results with ResNet50}}                                                                                                             \\

~TPT\textsuperscript{\dag}    & 60.7 & 40.8 & 28.3 & 17.6 & 62.7 & 74.9 & \underline{84.5} & 61.5 & 58.5 & 87.0 &-- &-- & --& \textbf{69.8} & \textit{(58.8)} \\
~CLIP-DN & 60.2 & 41.2 & 28.3 & 17.2 & 63.3 & 77.2 & 83.3 & 60.8 & 54.9 & \textbf{88.3} & 74.0 & 44.7 & 48.9 & 60.4 & 57.3 \\ 
~CLIP    & 60.3 & 41.1 & 26.9 & 16.7 & 62.9 & 76.7 & 83.1 & 61.2 & 54.5 & \underline{87.9} & 72.3 & 42.5 & 47.0 & 59.9 & 56.6 \\
\hspace{0.5em} + \textbf{\ours} & 62.2 & 42.8 & 31.9 & 17.4 & \underline{69.3} & 77.9 & 80.3 & 61.8 & 56.4 & 86.9 & 76.3 & 46.0 & 49.7 & 62.8 & 58.7 \\
\quad \emph{(vs CLIP)} & \gain{1.9} & \gain{1.7} & \gain{5.0} & \gain{0.7} & \gain{6.4} & \gain{1.2} & \loss{2.8} & \gain{0.6} & \gain{1.9} & \loss{1.0} & \gain{4.0} & \gain{3.5} & \gain{2.7} & \gain{2.9} & \gain{2.1} \\
\hspace{0.5em} + \textbf{\ourssparse} & \underline{62.9} & 43.1 & \textbf{38.8} & 17.9 & 68.8 & 78.3 & 77.7 & 61.2 & 55.8 & 86.3 & 78.6 & 48.0 & 51.1 & 64.2 & 59.5 \\
\quad \emph{(vs CLIP)} & \gain{2.6} & \gain{2.0} & \gain{11.9} & \gain{1.2} & \gain{5.9} & \gain{1.6} & \loss{5.4} & \gain{0.0} & \gain{1.3} & \loss{1.6} & \gain{6.3} & \gain{5.5} & \gain{4.1} & \gain{4.3} & \gain{2.9} \\
~InMaP   & \underline{62.9} & 45.7 & 33.6 & \textbf{19.2} & 66.4 & \textbf{79.2} & \textbf{85.7} & \underline{65.0} & \textbf{62.0} & 76.2 & 79.0 & 49.7 & \textbf{55.4} & 66.0 & 60.4 \\
\hspace{0.5em} + \textbf{\ours} & \underline{62.9} & \textbf{46.6} & \underline{36.3} & 18.7 & 69.1 & \underline{79.0} & 83.4 & 64.9 & 61.8 & 80.2 & \underline{79.1} & \textbf{50.6} & 54.9 & 66.5 & \textbf{61.0} \\
\quad \emph{(vs CLIP)} & \gain{2.6} & \gain{5.5} & \gain{9.4} & \gain{2.0} & \gain{6.2} & \gain{2.3} & \gain{0.3} & \gain{3.7} & \gain{7.3} & \loss{7.7} & \gain{6.8} & \gain{8.1} & \gain{7.9} & \gain{6.6} & \gain{4.4} \\
\quad \emph{(vs InMaP)} & \gain{0.0} & \gain{0.9} & \gain{2.7} & \loss{0.5} & \gain{2.7} & \loss{0.2} & \loss{2.3} & \loss{0.1} & \loss{0.2} & \gain{4.0} & \gain{0.1} & \gain{0.9} & \loss{0.5} & \gain{0.5} & \gain{0.6} \\
\hspace{0.5em} + \textbf{\ourssparse} & \textbf{63.0} & \underline{46.3} & 36.2 & \underline{18.9} & \textbf{69.4} & \textbf{79.2} & 81.4 & \textbf{65.1} & \underline{61.9} & 79.3 & \textbf{79.2} & \underline{50.5} & \underline{55.1} & \underline{67.0} & \underline{60.9} \\
\quad \emph{(vs CLIP)} & \gain{2.7} & \gain{5.2} & \gain{9.3} & \gain{2.2} & \gain{6.5} & \gain{2.5} & \loss{1.7} & \gain{3.9} & \gain{7.4} & \loss{8.6} & \gain{6.9} & \gain{8.0} & \gain{8.1} & \gain{7.1} & \gain{4.3} \\
\quad \emph{(vs InMaP)} & \gain{0.1} & \gain{0.6} & \gain{2.6} & \loss{0.3} & \gain{3.0} & \gain{0.0} & \loss{4.3} & \gain{0.1} & \loss{0.1} & \gain{3.1} & \gain{0.2} & \gain{0.8} & \loss{0.3} & \gain{1.0} & \gain{0.5} \\
\midrule
\multicolumn{16}{l}{\textit{Results with ViT-B/16}} \\
~TPT\textsuperscript{\dag}    & 69.0 & 47.8 & 42.4 & 24.8 & 69.0 & 84.7 & 87.8 & 65.5 & 66.9 & \textbf{94.2} &-- &-- &-- & 68.0 & \textit{(65.5)} \\
~CLIP-DN & 68.3 & 45.6 & 53.3 & 24.3 & 67.9 & 86.0 & 87.7 & 66.5 & 64.1 & 93.6 & 91.5 & 69.6 & 56.0 & 68.4 & 67.3 \\
~CLIP    & 68.8 & 45.1 & 50.2 & 23.0 & 67.0 & 85.7 & 88.3 & 66.3 & 63.8 & \underline{93.9} & 91.2 & 68.7 & 55.2 & 67.5 & 66.8 \\
\hspace{0.5em} + \textbf{\ours} & 70.2 & 48.6 & 55.6 & 25.4 & 73.5 & 86.9 & 87.1 & 67.4 & 65.6 & 93.1 & 92.2 & 71.0 & 59.4 & 71.5 & 69.1 \\
\quad \emph{(vs CLIP)} & \gain{1.4} & \gain{3.5} & \gain{5.4} & \gain{2.4} & \gain{6.5} & \gain{1.2} & \loss{1.2} & \gain{1.1} & \gain{1.8} & \loss{0.8} & \gain{1.0} & \gain{2.3} & \gain{4.2} & \gain{4.0} & \gain{2.3} \\
\hspace{0.5em} + \textbf{\ourssparse} & 71.0 & 49.1 & 58.2 & 25.8 & 72.6 & 87.3 & 86.3 & 67.2 & 66.1 & 92.1 & 92.7 & 72.0 & 59.1 & 72.2 & 69.4 \\
\quad \emph{(vs CLIP)} & \gain{2.2} & \gain{4.0} & \gain{8.0} & \gain{2.8} & \gain{5.6} & \gain{1.6} & \loss{2.0} & \gain{0.9} & \gain{2.3} & \loss{1.8} & \gain{1.5} & \gain{3.3} & \gain{3.9} & \gain{4.7} & \gain{2.6} \\
~InMaP   & \underline{72.0} & 49.6 & 59.4 & \underline{29.0} & 71.9 & \textbf{87.9} & \textbf{91.6} & \textbf{71.4} & \textbf{71.9} & 79.0 & \underline{93.4} & 73.7 & 63.9 & \underline{75.4} & 70.7 \\
\hspace{0.5em} + \textbf{\ours} & \textbf{72.1} & \textbf{51.2} & \textbf{63.2} & \textbf{29.1} & \textbf{75.9} & \underline{87.8} & \underline{90.0} & \underline{71.0} & 71.2 & 84.0 & \underline{93.4} & \underline{74.0} & \textbf{64.3} & \textbf{76.3} & \textbf{71.7} \\
\quad \emph{(vs CLIP)} & \gain{3.3} & \gain{6.1} & \gain{13.0} & \gain{6.1} & \gain{8.9} & \gain{2.1} & \gain{1.7} & \gain{4.7} & \gain{7.4} & \loss{9.9} & \gain{2.2} & \gain{5.3} & \gain{9.1} & \gain{8.8} & \gain{4.9} \\
\quad \emph{(vs InMaP)} & \gain{0.1} & \gain{1.6} & \gain{3.8} & \gain{0.1} & \gain{4.0} & \loss{0.1} & \loss{1.6} & \loss{0.4} & \loss{0.7} & \gain{5.0} & \gain{0.0} & \gain{0.3} & \gain{0.4} & \gain{0.9} & \gain{1.0} \\
\hspace{0.5em} + \textbf{\ourssparse} & \textbf{72.1} & \underline{51.0} & \underline{62.7} & \underline{29.0} & \underline{75.5} & \textbf{87.9} & 89.0 & \textbf{71.4} & \underline{71.8} & 83.1 & \textbf{93.6} & \textbf{74.2} & \underline{64.2} & \textbf{76.3} & \underline{71.6} \\
\quad \emph{(vs CLIP)} & \gain{3.3} & \gain{5.9} & \gain{12.5} & \gain{6.0} & \gain{8.5} & \gain{2.2} & \gain{0.7} & \gain{5.1} & \gain{8.0} & \loss{10.8} & \gain{2.4} & \gain{5.5} & \gain{9.0} & \gain{8.8} & \gain{4.8} \\
\quad \emph{(vs InMaP)} & \gain{0.1} & \gain{1.4} & \gain{3.3} & \gain{0.0} & \gain{3.6} & \gain{0.0} & \loss{2.6} & \gain{0.0} & \loss{0.1} & \gain{4.1} & \gain{0.2} & \gain{0.5} & \gain{0.3} & \gain{0.9} & \gain{0.9} \\
\bottomrule
\end{tabular}
}

     \caption{\textbf{Inductive zero-shot classification accuracy on 14 datasets} for two CLIP backbones.
    Rows denoted as \emph{(vs CLIP)} and \emph{(vs InMaP)} show the absolute accuracy gains of our method over CLIP and InMaP, respectively.
      * denotes our method with approximation of $\hat{Y}$.
    \textsuperscript{\dag} denotes numbers taken from InMaP~\cite{qxh23}.
    }
    \label{tab:inductive}
\end{table*}
\paragraph{Leveraging LLM generated prompts}

In the main paper, we present the average results when we leverage LLM generated prompts from CuPL~\cite{plf22}. In Tables~\ref{tab:cupl_transductive} and~\ref{tab:cupl_inductive}, we present per dataset results for transductive and inductive setups, respectively. 
CuPL prompts, compared to hand-crafted universal class templates,  improve CLIP+\ours from 60.0\% to 64.6\% and from 58.7\% to 64.2\% for the transductive and inductive setup, respectively.

\begin{table*}
    \centering
    {
\footnotesize
\renewcommand*{\arraystretch}{1.2}
\begin{tabular}{lp{.58cm}p{.58cm}p{.58cm}p{.58cm}p{.58cm}p{.58cm}p{.58cm}p{.58cm}p{.58cm}p{.58cm}p{.58cm}p{.58cm}|p{.58cm}}
\toprule
        & \scriptsize  \rotatebox{30}{imagenet} & \scriptsize  \rotatebox{30}{dtd} & \scriptsize  \rotatebox{30}{fgvca} & \scriptsize  \rotatebox{30}{flowers} & \scriptsize  \rotatebox{30}{food} & \scriptsize  \rotatebox{30}{pets} & \scriptsize  \rotatebox{30}{sun} & \scriptsize  \rotatebox{30}{cars} & \scriptsize  \rotatebox{30}{caltech} & \scriptsize  \rotatebox{30}{cifar10} & \scriptsize  \rotatebox{30}{cifar100} & \scriptsize  \rotatebox{30}{ucf} & \scriptsize  \rotatebox{30}{avg} \\
\midrule
\multicolumn{14}{l}{\textit{Results with ResNet50}}                                                                                                             \\
~CLIP    & 61.7 & 49.1 & 18.5 & \underline{67.9} & 77.8 & 87.5 & 63.8 & 55.8 & \underline{88.7} & 76.4 & 45.2 & 63.5 & 63.0 \\
\hspace{0.5em} + \textbf{\ours} & 62.7 & 51.4 & \underline{20.2} & 67.6 & 78.9 & 88.1 & 65.2 & 58.8 & \textbf{89.8} & 77.6 & 47.4 & 67.8 & 64.6 \\
\quad \emph{(vs CLIP)} & \gain{1.0} & \gain{2.3} & \gain{1.7} & \loss{0.3} & \gain{1.1} & \gain{0.6} & \gain{1.4} & \gain{3.0} & \gain{1.1} & \gain{1.2} & \gain{2.2} & \gain{4.3} & \gain{1.6} \\
~InMaP   & \textbf{64.4} & \underline{54.5} & \textbf{22.2} & 67.2 & \textbf{79.3} & \textbf{89.9} & \underline{67.4} & \underline{62.8} & 73.7 & \underline{78.2} & \underline{50.2} & \underline{68.2} & \underline{64.8} \\ 
\hspace{0.5em} + \textbf{\ours} & \underline{64.3} & \textbf{55.6} & \textbf{22.2} & \textbf{69.8} & \underline{79.2} & \underline{89.5} & \textbf{67.8} & \textbf{63.2} & 78.9 & \textbf{78.9} & \textbf{50.5} & \textbf{70.2} & \textbf{65.8} \\
\quad \emph{(vs CLIP)} & \gain{2.6} & \gain{6.5} & \gain{3.7} & \gain{1.9} & \gain{1.4} & \gain{2.0} & \gain{4.0} & \gain{7.4} & \loss{9.8} & \gain{2.5} & \gain{5.3} & \gain{6.7} & \gain{2.8} \\
\quad \emph{(vs InMaP)} & \loss{0.1} & \gain{1.1} & \gain{0.0} & \gain{2.6} & \loss{0.1} & \loss{0.4} & \gain{0.4} & \gain{0.4} & \gain{5.2} & \gain{0.7} & \gain{0.3} & \gain{2.0} & \gain{1.0} \\
\midrule
\multicolumn{14}{l}{\textit{Results with ViT-B/16}} \\
~CLIP    & 70.0 & 53.2 & 27.9 & 73.4 & 86.3 & 91.7 & 69.5 & 66.1 & \textbf{94.4} & 90.7 & 69.4 & 70.5 & 71.9 \\
\hspace{0.5em} + \textbf{\ours} & \underline{70.5} & 54.0 & 30.1 & 72.2 & 86.9 & 91.8 & 69.7 & \underline{67.3} & \underline{92.7} & 92.4 & 69.9 & 74.0 & 72.6 \\ 
\quad \emph{(vs CLIP)} & \gain{0.5} & \gain{0.8} & \gain{2.2} & \loss{1.2} & \gain{0.6} & \gain{0.1} & \gain{0.2} & \gain{1.2} & \loss{1.7} & \gain{1.7} & \gain{0.5} & \gain{3.5} & \gain{0.7} \\
~InMaP   & \textbf{73.3} & \underline{57.3} & \textbf{31.9} & \underline{74.1} & \textbf{88.1} & \textbf{93.7} & \underline{73.3} & \textbf{72.8} & 78.0 & \underline{93.4} & \textbf{73.3} & \underline{77.1} & 73.9 \\
\hspace{0.5em} + \textbf{\ours} & \textbf{73.3} & \textbf{57.9} & \underline{31.7} & \textbf{76.9} & \underline{88.0} & \underline{93.3} & \textbf{73.7} & \textbf{72.8} & 83.3 & \textbf{93.6} & \underline{73.2} & \textbf{79.5} & \textbf{74.8} \\
\quad \emph{(vs CLIP)} & \gain{3.3} & \gain{4.7} & \gain{3.8} & \gain{3.5} & \gain{1.7} & \gain{1.6} & \gain{4.2} & \gain{6.7} & \loss{11.1} & \gain{2.9} & \gain{3.8} & \gain{9.0} & \gain{2.9} \\
\quad \emph{(vs InMaP)} & \gain{0.0} & \gain{0.6} & \loss{0.2} & \gain{2.8} & \loss{0.1} & \loss{0.4} & \gain{0.4} & \gain{0.0} & \gain{5.3} & \gain{0.2} & \loss{0.1} & \gain{2.4} & \gain{0.9} \\
\bottomrule
\end{tabular}
}
    \caption{\textbf{Transductive zero-shot classification accuracy on 12 datasets} for two CLIP backbones and prompts generated by a LLM~\cite{plf22}. 
    Rows denoted as \emph{(vs CLIP)} and \emph{(vs InMaP)} show the absolute accuracy gains of our method over CLIP and InMaP, respectively.
    }
    \label{tab:cupl_transductive}
\end{table*}

\begin{table*}
    \centering
    {
\footnotesize
\renewcommand*{\arraystretch}{1.2}
\begin{tabular}{lp{.58cm}p{.58cm}p{.58cm}p{.58cm}p{.58cm}p{.58cm}p{.58cm}p{.58cm}p{.58cm}p{.58cm}p{.58cm}p{.58cm}|p{.58cm}}
\toprule
        & \scriptsize  \rotatebox{30}{imagenet} & \scriptsize  \rotatebox{30}{dtd} & \scriptsize  \rotatebox{30}{fgvca} & \scriptsize  \rotatebox{30}{flowers} & \scriptsize  \rotatebox{30}{food} & \scriptsize  \rotatebox{30}{pets} & \scriptsize  \rotatebox{30}{sun} & \scriptsize  \rotatebox{30}{cars} & \scriptsize  \rotatebox{30}{caltech} & \scriptsize  \rotatebox{30}{cifar10} & \scriptsize  \rotatebox{30}{cifar100} & \scriptsize  \rotatebox{30}{ucf} & \scriptsize  \rotatebox{30}{avg} \\
\midrule
\multicolumn{14}{l}{\textit{Results with ResNet50}}                                                                                                             \\
~CLIP    & 61.7 & 49.1 & 18.5 & 67.9 & 77.8 & \textbf{87.5} & 63.8 & 55.8 & \textbf{88.7} & 76.4 & 45.2 & 63.5 & 63.0 \\
\hspace{0.5em} + \textbf{\ours} & \underline{63.1} & 51.4 & \underline{20.0} & \textbf{72.7} & \underline{78.4} & 85.4 & 63.3 & 57.8 & \underline{88.3} & 77.9 & 48.0 & 63.6 & 64.2 \\
\quad \emph{(vs CLIP)} & \gain{1.4} & \gain{2.3} & \gain{1.5} & \gain{4.8} & \gain{0.6} & \loss{2.1} & \loss{0.5} & \gain{2.0} & \loss{0.4} & \gain{1.5} & \gain{2.8} & \gain{0.1} & \gain{1.2} \\
~InMaP   & \textbf{63.4} & \textbf{54.6} & \textbf{22.6} & 68.8 & \textbf{79.1} & \underline{86.4} & \textbf{66.6} & \textbf{62.5} & 75.7 & \underline{78.2} & \underline{50.4} & \underline{67.5} & \underline{64.6} \\
\hspace{0.5em} + \textbf{\ours} & \textbf{63.4} & \underline{54.1} & \textbf{22.6} & \underline{71.5} & \textbf{79.1} & 83.4 & \underline{66.5} & \underline{62.4} & 79.4 & \textbf{78.8} & \textbf{51.0} & \textbf{67.7} & \textbf{65.0} \\
\quad \emph{(vs CLIP)} & \gain{1.7} & \gain{5.0} & \gain{4.1} & \gain{3.6} & \gain{1.3} & \loss{4.1} & \gain{2.7} & \gain{6.6} & \loss{9.3} & \gain{2.4} & \gain{5.8} & \gain{4.2} & \gain{2.0} \\
\quad \emph{(vs InMaP)} & \gain{0.0} & \loss{0.5} & \gain{0.0} & \gain{2.7} & \gain{0.0} & \loss{3.0} & \loss{0.1} & \loss{0.1} & \gain{3.7} & \gain{0.6} & \gain{0.6} & \gain{0.2} & \gain{0.4} \\
\midrule
\multicolumn{14}{l}{\textit{Results with ViT-B/16}} \\
~CLIP    & 70.0 & 53.2 & 27.9 & 73.4 & 86.3 & \underline{91.7} & 69.5 & 66.1 & \textbf{94.4} & 90.7 & 69.4 & 70.5 & 71.9 \\
\hspace{0.5em} + \textbf{\ours} & 71.2 & 55.5 & \underline{29.8} & \underline{77.7} & 87.2 & 91.1 & 69.7 & \underline{67.5} & \textbf{94.4} & 91.6 & 71.3 & \underline{72.6} & 73.3 \\
\quad \emph{(vs CLIP)} & \gain{1.2} & \gain{2.3} & \gain{1.9} & \gain{4.3} & \gain{0.9} & \loss{0.6} & \gain{0.2} & \gain{1.4} & \gain{0.0} & \gain{0.9} & \gain{1.9} & \gain{2.1} & \gain{1.4} \\
~InMaP   & \underline{72.4} & \textbf{57.2} & \textbf{32.8} & 75.8 & \textbf{88.0} & \textbf{92.3} & \textbf{73.0} & \textbf{72.9} & 79.8 & \underline{93.3} & \underline{73.7} & \textbf{76.6} & \underline{74.0} \\
\hspace{0.5em} + \textbf{\ours} & \textbf{72.5} & \underline{56.3} & \textbf{32.8} & \textbf{78.5} & \underline{87.9} & 89.6 & \underline{72.6} & \textbf{72.9} & \underline{83.9} & \textbf{93.5} & \textbf{73.8} & \textbf{76.6} & \textbf{74.2} \\
\quad \emph{(vs CLIP)} & \gain{2.5} & \gain{3.1} & \gain{4.9} & \gain{5.1} & \gain{1.6} & \loss{2.1} & \gain{3.1} & \gain{6.8} & \loss{10.5} & \gain{2.8} & \gain{4.4} & \gain{6.1} & \gain{2.3} \\
\quad \emph{(vs InMaP)} & \gain{0.1} & \loss{0.9} & \gain{0.0} & \gain{2.7} & \loss{0.1} & \loss{2.7} & \loss{0.4} & \gain{0.0} & \gain{4.1} & \gain{0.2} & \gain{0.1} & \gain{0.0} & \gain{0.2} \\
\bottomrule
\end{tabular}
}
    \caption{\textbf{Inductive zero-shot classification accuracy on 12 datasets} for two CLIP backbones and prompts generated by a LLM~\cite{plf22}. 
    Rows denoted as \emph{(vs CLIP)} and \emph{(vs InMaP)} show the absolute accuracy gains of our method over CLIP and InMaP, respectively.
    }
    \label{tab:cupl_inductive}
\end{table*}
\paragraph{Web-crawled unlabeled images}

We construct a new set of unlabeled images with 10,000 images per class that are chosen either randomly, or based on proximity of their image or text features to the class representation. 
Results are presented in Table~\ref{tab:web}.
Switching to using only the LAION-based unlabeled set, we observe that random selection fails by performing worse than CLIP, but the other two options provide some improvement, with the caption-based neighbors being a bit better. Interestingly, web-crawling is better than the the target distribution images for the Pets dataset, while much worse for Eurosat due to the lack of satellite images on LAION. 
On the other hand, if the randomly selected set is mixed with that from the target distribution, \ours manages to benefit from the relevant images and to deliver an improvement compared to CLIP.

\begin{table*}[t]
    \centering
    \resizebox{\linewidth}{!}{
\renewcommand*{\arraystretch}{1.2}
\footnotesize
\begin{tabular}{lp{.58cm}p{.58cm}p{.58cm}p{.58cm}p{.58cm}p{.58cm}p{.58cm}p{.58cm}p{.58cm}p{.58cm}p{.58cm}p{.58cm}p{.58cm}p{.58cm}|p{.58cm}}
\toprule
        & \scriptsize  \rotatebox{30}{imagenet} & \scriptsize  \rotatebox{30}{dtd} & \scriptsize  \rotatebox{30}{eurosat} & \scriptsize  \rotatebox{30}{fgvca} & \scriptsize  \rotatebox{30}{flowers} & \scriptsize  \rotatebox{30}{food} & \scriptsize  \rotatebox{30}{pets} & \scriptsize  \rotatebox{30}{sun} & \scriptsize  \rotatebox{30}{cars} & \scriptsize  \rotatebox{30}{caltech} & \scriptsize  \rotatebox{30}{cifar10} & \scriptsize  \rotatebox{30}{cifar100} & \scriptsize  \rotatebox{30}{cub} & \scriptsize  \rotatebox{30}{ucf} & \scriptsize  \rotatebox{30}{avg} \\
\midrule
\multicolumn{16}{l}{\textit{Results with ResNet50}}                                                                                                             \\

~CLIP    & 60.3 & 41.1 & 26.9 & 16.7 & 62.9 & 76.7 & 83.1 & 61.2 & 54.5 & \underline{87.9} & 72.3 & 42.5 & 47.0 & 59.9 & 56.6 \\
\hspace{0.5em} + \textbf{\ours} (target distribution) & \textbf{62.2} & \textbf{42.8} & \textbf{31.9} & \textbf{17.4} & \textbf{69.3} & \textbf{77.9} & 80.3 & 61.8 & \textbf{56.4} & 86.9 & \textbf{76.3} & \textbf{46.0} & \textbf{49.7} & \textbf{62.8} & \textbf{58.7} \\
\hspace{0.5em} + \textbf{\ours} (target distr. + LAION random) & \underline{61.4} & \underline{42.3} & \underline{30.2} & 15.7 & \underline{63.5} & \underline{77.1} & 80.3 & 61.6 & 53.7 & 87.8 & \underline{75.1} & \underline{42.9} & 47.5 & 59.8 & \underline{57.1} \\
\hspace{0.5em} + \textbf{\ours} (LAION random) & 59.9 & 41.4 & 26.2 & 14.3 & 59.1 & 74.5 & 79.3 & 61.1 & 51.4 & 87.6 & 70.6 & 41.4 & 43.4 & 58.8 & 54.6 \\
\hspace{0.5em} + \textbf{\ours} (LAION image neighbors) & 60.6 & 41.1 & 29.1 & 16.7 & \underline{63.5} & 76.9 & \underline{83.5} & \underline{61.9} & 54.7 & \textbf{88.4} & 69.5 & 41.1 & 48.2 & 59.6 & 56.8 \\
\hspace{0.5em} + \textbf{\ours} (LAION caption neighbors) & 60.7 & 40.5 & 26.9 & \underline{16.9} & 63.0 & 76.9 & \textbf{83.6} & \textbf{62.0} & \underline{55.3} & \textbf{88.4} & 73.0 & 41.7 & \underline{48.5} & \underline{60.1} & 57.0 \\
\midrule
\multicolumn{16}{l}{\textit{Results with ViT-B/16}} \\
~CLIP    & 68.8 & 45.1 & 50.2 & 23.0 & 67.0 & 85.7 & \underline{88.3} & 66.3 & 63.8 & 93.9 & 91.2 & 68.7 & 55.2 & 67.5 & 66.8 \\
\hspace{0.5em} + \textbf{\ours} (target distribution) & \textbf{70.2} & \textbf{48.6} & \textbf{55.6} & \textbf{25.4} & \textbf{73.5} & \textbf{86.9} & 87.1 & \textbf{67.4} & \textbf{65.6} & 93.1 & \textbf{92.2} & \textbf{71.0} & \textbf{59.4} & \textbf{71.5} & \textbf{69.1} \\
\hspace{0.5em} + \textbf{\ours} (target distr.  + LAION random) & \underline{69.5} & \underline{45.9} & \underline{53.1} & 21.0 & 67.3 & \underline{86.3} & 86.4 & 66.9 & \underline{64.7} & 93.7 & \underline{91.9} & \underline{69.3} & 55.6 & \underline{67.6} & \underline{67.1} \\
\hspace{0.5em} + \textbf{\ours} (LAION random) & 68.6 & 44.9 & 49.4 & 19.8 & 65.3 & 85.3 & 86.8 & 66.5 & 63.0 & 93.6 & 90.3 & 68.6 & 54.0 & 66.9 & 65.9 \\
\hspace{0.5em} + \textbf{\ours} (LAION image neighbors) & 69.0 & 45.4 & 49.2 & \underline{23.8} & \underline{68.1} & 85.8 & \textbf{88.4} & 66.9 & 64.5 & \textbf{94.2} & 90.8 & 68.1 & \underline{56.7} & \underline{67.6} & 67.0 \\
\hspace{0.5em} + \textbf{\ours} (LAION  caption neighbors) & 69.1 & 45.0 & 49.4 & 23.4 & \underline{68.1} & 85.9 & \textbf{88.4} & \underline{67.0} & 64.6 & \underline{94.0} & 90.8 & 68.5 & \underline{56.7} & \underline{67.6} & \underline{67.1} \\
\bottomrule
\end{tabular}
}
    \caption{\textbf{Inductive zero-shot classification accuracy on 14 datasets using different sources of unlabeled data.} Compared to the original experiments that use unlabeled images from the target distribution, LAION-400M is used to create a web-crawled unlabeled set.
    }
    \label{tab:web}
\end{table*}

\end{document}